\documentclass{article}

\usepackage[preprint]{neurips_2025}

\usepackage[utf8]{inputenc} 
\usepackage[T1]{fontenc}    

\usepackage[colorlinks,
            linkcolor=blue,
            anchorcolor=blue,
            citecolor=blue,
            urlcolor=blue,
            ]{hyperref}       

\usepackage{url}            
\usepackage{booktabs}       
\usepackage{amsfonts}       
\usepackage{nicefrac}       
\usepackage{microtype}      
\usepackage{xcolor}         
\usepackage{algorithm}  
\usepackage{algorithmic}  
\usepackage{graphicx}
\usepackage{amsmath} 
\usepackage{array}      
\usepackage{booktabs}   
\usepackage{subfigure}
\usepackage{xcolor}
\usepackage{multirow}

\usepackage{wrapfig}

\usepackage{makecell}

\usepackage[most]{tcolorbox}
\newtcbox{\mybox}[1][red]{on line, arc = 0pt, outer arc = 0pt, colback = #1!10!white, colframe = #1!50!black, boxsep = 0pt, left = 1pt, right = 1pt, top = 2pt, bottom = 2pt, boxrule = 0pt, bottomrule = 1pt, toprule = 1pt}
\usepackage{enumitem}

\usepackage{amsmath}
\usepackage{amssymb}
\usepackage{mathtools}
\usepackage{amsthm}

\title{From Macro to Micro: Probing Dataset Diversity in Language Model Fine-Tuning}

\author{%
  Haoyu Li \textsuperscript{1, *}, 
  Xuhong Li \textsuperscript{2, *}, 
  Yiming Dong \textsuperscript{3}, 
  Kun Liu \textsuperscript{1, \dag} 
  \\
  \small 1. School of Automation, Beijing Institute of Technology \\
  \small 2. Baidu Inc. \\
  \small 3. School of Physics, Peking University \\
  \texttt{haoyli\_bit@bit.edu.cn}, 
  \texttt{lixuhong@baidu.com}, \\
  \texttt{ydong@pku.edu.cn},  
  \texttt{kunliubit@bit.edu.cn} \\
}

\begin{document}

\maketitle

\vspace{-0.2cm}
\begin{abstract}

Dataset diversity plays a pivotal role for the successful training of many machine learning models, particularly in the supervised fine-tuning (SFT) stage of large language model (LLM) development. 
Despite increasing recognition of its importance, systematic analyses of dataset diversity still remain underexplored. 
To address this gap, this work presents a systematic taxonomy of existing diversity-control strategies, which primarily focus on the \textit{instruction} component, operating at either \underline{macroscopic} (entire instruction semantics) or \underline{mesoscopic} levels (instruction units), and furthermore introduces a novel analysis of \underline{microscopic} diversity within the \textit{response} component, specifically analyzing the statistical distribution of \textbf{tokens} in SFT training samples.
In the experimental evaluation, we construct fixed-size datasets (e.g., 10,000 samples each) from a corpus of 117,000 open-source SFT samples, incorporating six distinct diversity-control strategies spanning macro-, meso-, and microscopic levels applied to both instructions and responses. 
We then fine-tune LLMs on these datasets to assess the six diversity-control strategies.
Results reveal that while macroscopic and mesoscopic strategies lead to higher performance with increasing diversity, the microscopic strategy in responses exhibits both a stronger correlation between model performance and the degree of diversity and superior performance with maximum diversity across all strategies.
These findings offer actionable insights for constructing high-performance SFT datasets.
\end{abstract}
\footnotetext[1]{Equal contribution: Work done during Haoyu Li's internship at Baidu Inc.}
\footnotetext[2]{Corresponding author}
\vspace{-0.2cm}
\section{Introduction}
\vspace{-0.1cm}
The success of large language models (LLMs) hinges not only on the advancements in model architectures and computational resources, but also critically on the acquisition and management of training data~\cite{kaplan2020scaling,ouyang2022training,achiam2023gpt,wang2023self}.
While significant research has focused on the quality of training samples, increasing attention is being given to dataset diversity that enhances models' ability to generalize and handle real-world scenarios~\cite{bubeck2023sparks,bukharin2023QDIT,pmlr-v235-zhao24a,zhou2024lima}.
This work advances this line of research and focuses on the dataset diversity of the supervised fine-tuning (SFT) stage in the LLM alignment~\cite{ouyang2022training}. 

Existing diversity-control strategies prove to be effective in enhancing the LLM capacity.
The first strategy~\cite{grootendorst2022bertopic,du2023mods,ge-etal-2024-car,ge2024scaling} is to cluster instructions by considering the entire semantics across multiple data sources (including both open-sourced datasets and proprietary user prompts), and improve the dataset diversity by utilizing more clusters as possible.
An alternative strategy, exemplified by InsTag~\cite{lu2023instag}, further decomposes the instruction into atomic instruction components, and increases the dataset diversity by covering more instruction unit tags.

While these strategies prove to be effective, there are two challenges along this research direction.
First, systematic analyses and metrics of dataset diversity still remain underexplored.
Different datasets, models and evaluations are used for experiments in previous works, and few metrics can correlate dataset diversity to the model performance significantly.
Secondly, previous works focus on the \textit{instruction} component of the instruction-response pair in the SFT dataset.
Note that LLMs are conventionally supervised using \textit{responses} as the primary training signals~\cite{brown2020gpt3,hu2022lora,dettmers2024qlora}.
Though instructions serve as indicators of the diversity of topics, domains, disciplines or other semantic aspects~\cite{wei2022emergent,wang2023self,zhou2024lima} and may implicitly diversify responses, explicit signals might be more effective.

\begin{figure*}[t]
\centering 
\includegraphics[width=0.95\textwidth]{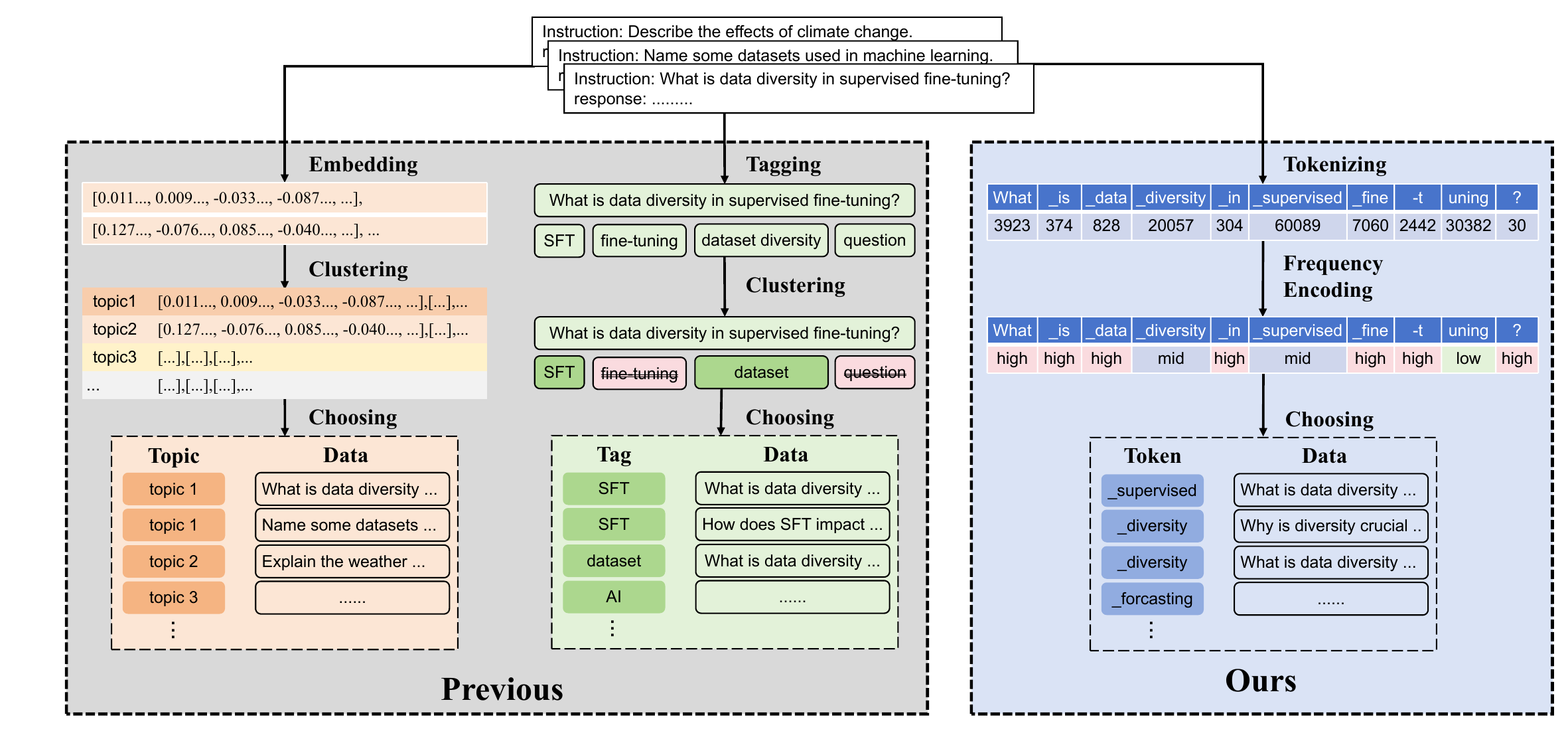}
\vspace{-0.2cm}
\caption{Diversity-control strategies across three scales on instruction. At the macroscopic scale, each instruction is assigned to a corresponding topic after embedding and clustering~\cite{grootendorst2022bertopic}. At the mesoscopic scale, each instruction is linked to multiple relevant tags through LLM-based tagging, filtering, and clustering~\cite{lu2023instag}. At the microscopic scale, we tokenize all instructions and select representative tokens based on their frequencies across the corpus for each instruction. Diversity-controlled datasets are then constructed based on topic/tag/token varieties, which also serve as diversity indicators. The same diversity-control strategies are also applied to the \textbf{response} perspective.}
\label{fig:three-level-way}
\vspace{-0.3cm} 
\end{figure*}





To tackle the above challenges, we first present a taxonomy on the diversity strategies from macro- (entire instructions), meso- (instruction units) and microscopic (tokens) levels and on both instruction and response components during the SFT stage of LLM training.
We then propose to explicitly examine the impact of controlling diversity within the \textit{response} component of SFT datasets and compare its effectiveness to that of the instruction component.
For comprehensive comparisons, we apply various diversity control strategies to construct SFT datasets from 117K open-source instructions where the responses are re-constructed for quality control, and train hundreds of models using Llama series models~\cite{2023llama2,dubey2024llama3} for quantifying the effectiveness of different diversity strategies.

Despite these efforts, defining dataset diversity and establishing robust metrics correlating diversity with model performance remain challenging.
We collect multiple diversity-related metrics and involve an additional metric relating to information theory, which may serve as a powerful tool for quantifying dataset characteristics.
We present analyses and discuss the advantages and limitations of these metrics.
Specifically, the contributions of this work can be summarized as follows:
\vspace{-0.1cm}

{\textbf{(1)}} A systematic taxonomy on dataset diversity-control strategies is proposed, from macro-, meso- to microscopic levels on instruction and response components of SFT datasets. 
\vspace{-0.1cm}

{\textbf{(2)}} A novel proposed approach on controlling the response diversity from the token (microscopic) level exhibits both a stronger correlation between model performance and the degree of diversity and superior performance with maximum diversity across all strategies. 
\vspace{-0.1cm}

{\textbf{(3)}} Comprehensive experiments are conducted involving fix-size datasets, incorporating six distinct diversity-control strategies (spanning three levels, applied to both instructions and responses). Detailed analyses across three dataset sizes, three strategies, two components, multiple diversity metrics, are also provided. All results demonstrate that macroscopic and mesoscopic strategies effectively enhance model performance with increasing diversity, while the microscopic strategy on responses not only displays stronger performance-diversity correlation, but also achieves the optimal model performance among all strategies when diversity is maximized.
\vspace{-0.1cm}

\section{Methodology and Framework}
\label{sec:chapter2}
\vspace{-0.1cm}

In this section, we introduce the three-level perspective of dataset diversification strategies.
We first review the macro- and mesoscopic ones on the instruction component in Section~\ref{subsec:macro-and-meso}.
Then we present the proposed strategy of the microscopic one on the response component in Section~\ref{subsec:micro}.
The illustration of previous works and our proposed method is shown in Figure~\ref{fig:three-level-way}.
We eventually introduce the framework of empirical comparisons and analyses across all strategy combinations, including data source, dataset construction, model training evaluations and the diversity metrics, in Section~\ref{subsec:framework}.

\vspace{-0.1cm}
\subsection{Previous Works: Macro- and Mesoscopic Diversity Strategies}
\label{subsec:macro-and-meso}

\textbf{Macroscopic} analysis of dataset diversity focuses on the semantic content of texts to characterize thematic diversity. 
BERTopic~\cite{grootendorst2022bertopic} is one typical approach of this strategy and employs the following steps:
(1) Embedding the texts into dense vector representations via any tokenizer (e.g.,~\cite{mikolov2013word2vec}); 
(2) Clustering embeddings to semantically related clusters (e.g., HDBSCAN~\cite{ester1996dbscan,mcinnes2017hdbscan}), where dimension-reduction techniques may be helpful (e.g., UMAP~\cite{mcinnes2018umap}).
Then the dataset diversity scale can be controlled by choosing samples from a certain amount of clusters.

Multiple approaches~\cite{du2023mods,ge-etal-2024-car,ge2024scaling} can be categorized into this macroscopic strategy with various differences.
For example, CaR~\cite{ge-etal-2024-car} additionally ranks the samples within each cluster and selects the top-quality samples.
MoDS~\cite{du2023mods} computes the embeddings and selects examples by maximizing the embedding distance.
A persona-based method~\cite{ge2024scaling} has been proposed to create synthetic instructions and maximize the diversity of instructions.

\textbf{Mesoscopic} analysis of dataset diversity focuses on decomposing an entire text into several unit tags, and then clustering the tags instead of the entire semantic content of texts.
InsTag~\cite{lu2023instag} is the typical approach, which is achieved by the two-step processing approach:
(1) Generating unit tags for each text via an LLM;
(2) Clustering tags into semantically coherent groups.
Similar to macroscopic strategy, the dataset diversity can be also controlled by including different amounts of tags.

\subsection{Our Proposed Method: The Microscopic Strategy on Responses}
\label{subsec:micro}

\begin{wrapfigure}{r}{0.5\textwidth}
\centering 
\vspace{-0.2cm}
\includegraphics[width=0.5\columnwidth]{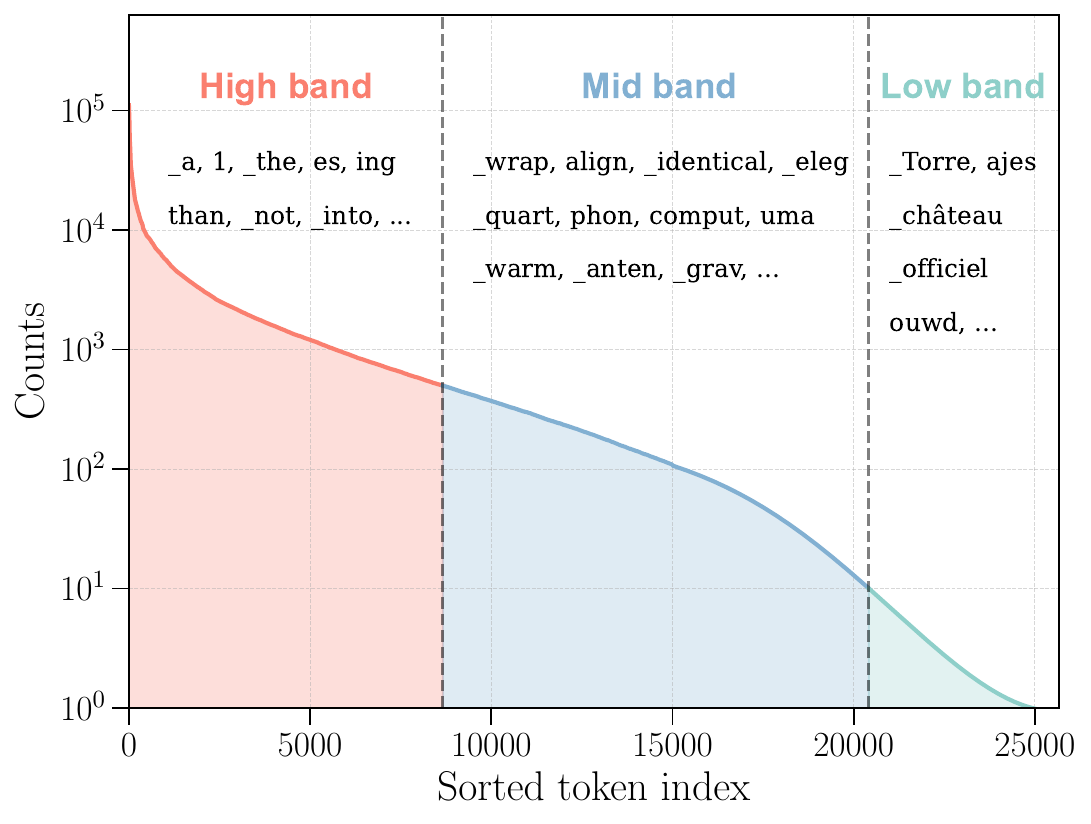}
\vspace{-0.2cm}
\caption{Distribution for tokens in the SFT dataset. Based on the counts of tokens, we classify them into three categories: ``High band'' (more than 500 counts), ``Mid band'' (10-500 counts), and ``Low band'' (fewer than 10 counts).} 
\label{token-dis}
\vspace{-0.4cm} 
\end{wrapfigure}

While macro- and mesoscopic analyses provide valuable insights into the dataset diversity, existing approaches predominantly concentrate on \textit{the instruction component}.
Although instructions act as proxies for contextual diversity, they are not an explicit signal as the LLMs are supervised on responses only in convention during the SFT stage.
This fundamental misalignment motivates our dual innovation: (1) a paradigm shift from instruction-centric to response-driven diversity analysis, and (2) the introduction of a microscopic characterization framework operating at the token granularity. 
Specifically, the microscopic strategy on \textit{the response component} involves the following two steps:

\textbf{(1) Tokenizing Texts.}
The proposed method begins by tokenizing all responses (or instructions, without loss of generality) using an LLM's tokenizer. 
Token frequency, defined as \textit{the number of occurrences of a token that appears in the training samples}, is then calculated.
A manual inspection of tokens across the frequency spectrum leads to their classification into three bands: high, mid, and low (Figure~\ref{token-dis}).
High-band tokens are typically prepositions, articles, letters, numbers, and common prefixes and suffixes, which often lack specific semantic meaning. 
Conversely, low-band tokens tend to be names, loanwords, or typos. 
Consequently, \textit{mid-band tokens} carry most of the semantic meaning for the texts, and are flagged for subsequent processing.

\textbf{(2) Controlling the Dataset Diversity at Token-Level.}
Controlling the diversity at the token level is more challenging than the query or tag levels. One sample may contain over 20 mid-band tokens in average, compared to one semantic cluster or four tags.
This quantitative complexity leads to a non-uniform dataset distribution, causing a problem that cannot be trivially resolved by sampling samples from a predefined number of clusters or tags, as previous works do, because the number of token types cannot be controlled in such way.
We therefore propose a two-stage algorithm to control the number of clusters and samples separately.
Specifically, given a controlled diversity scale (the number of token types), we first prune the entire dataset to control the number of tokens using an inverse greedy pruning Algorithm~\ref{alg:stage0}.
This algorithm yields a moderate dataset that reaches the predefined dataset diversity scale yet with some redundant samples.
Then, we apply a token-aware sampling Algorithm~\ref{alg:stage1} to reduce redundant samples while keeping the diversity scale.

\subsection{Comprehensive Framework of Comparison}
\label{subsec:framework}

Previous dataset diversity methods are designed to aim at the instruction component while they can be flexibly applied to either the instruction or response component.
To systematically compare previous diversity strategies and our proposed approach, we propose a taxonomy and a framework to contain a coherent dataset construction, model training and evaluation pipeline with the dataset diversity strategy modifiable.
Within the proposed framework, we conduct experiments to validate the effectiveness of our proposed approach and advance this line of research.



\textbf{Training Data Preparation.}
We compile a comprehensive dataset from a variety of open-source resources.
After a rigorous cleaning process, which includes deduplication and filtering based on character encoding, a refined instructional set of 117K prompts is left.
To alleviate the quality variance, we ignore the original responses and reconstruct the paired responses by prompting Llama-3.1-70B-Nemotron model\footnote{It is the best open-source model on \url{lmarena.ai} at the time when starting this work.}~\cite{wang2024Nemotron}, to ensure that individual training samples are at a similar level of quality, without forgetting that the decoding strategy still involves some variance.
Dataset resources used in this work are documented in Appendix~\ref{subsec:dataset}.

From the same data source, we employ a diversity strategy to construct a series of SFT datasets spanning a spectrum from minimal to maximal diversity under this strategy. We maintain fixed dataset sizes across experiments while exploring three distinct scales: 10K, 20K, and 30K samples per experimental set.
All dataset construction strategies in this work culminate in a uniform selection process, resulting in datasets formally defined as: $\mathcal{D}_k = \bigcup_{i=1}^{k} \Pi(\mathcal{C}_i)$, 
where $k$ is the number of topic/tag/token types controlling the diversity scale, $\mathcal{C}_i$ refers to the set of samples regarding to the $i$-th type, and $\Pi$ represents the sampling strategy designed to achieve a uniform distribution across target types in $\mathcal{D}_k$.
This generates a series of datasets for each diversity strategy denoted as $\{\mathcal{D}_m\}_{m=k_1, k_2, \cdots, k_M}$, where $k_1$ and $k_M$ specify the minimal and maximal diversity bounds, $M$ indicates the number of datasets in the series (typically set to 7 in our experiments), and intermediate $k_i$ values can be algorithmically determined.

\begin{table*}[h]
\centering

\renewcommand\arraystretch{1.05}
\caption{Posterior diversity metrics. The upward-pointing arrow denotes a positive correlation between the metric value and dataset diversity, while the downward-pointing arrow indicates an inverse relationship (negative correlation).}
\vspace{0.1cm}
\resizebox{\textwidth}{!}{%
\begin{tabular}{l m{4cm} m{6cm}}
\toprule
Metric & Equation & Explanation \\
\midrule

N-gram Ratio (NR) $\uparrow$ & \( R_n = \frac{\# \text{Unique } n\text{-grams}}{\# \text{Total } n\text{-grams}} \) & The ratio of unique \( n \)-grams to total \( n \)-grams, serving as a measure of lexical diversity~\cite{padmakumar2024does_ngramdiv}. 
\\[0.4cm]

Embedding Distance (ED) $\uparrow$ & \( D_{\text{avg}} = \frac{1}{N} \sum_{a \neq b} \|\mathbf{v}_a - \mathbf{v}_b\| \) & The average distance between embeddings~\cite{arora2017embeddingdis}, where $\bf{v}$ is the embedding of an element.
\\[0.4cm]

Sequence Length (SL) $\uparrow$ & \( L_{\text{seq}} = \frac{1}{N} \sum_{i=1}^{N} l_i \) & The mean number of tokens over sequences~\cite{zhao2024longlength}.
\\[0.4cm]

Compression Ratio (CR) $\uparrow$ & \( C_{\text{ratio}} = \frac{\text{Original size}}{\text{Compressed size}} \) & The ratio of original dataset size to compressed size~\cite{shaib2024standardizing}.
\\[0.4cm]

Self-BLEU (BL) $\downarrow$ & \( S_{\text{BLEU}} = \frac{1}{N} \sum_{i=1}^{N} \text{BLEU}_{-i} \) & The average $BLEU_{-i}$ score where each text $i$ is compared against the rest of the dataset ($-i$)~\cite{zhu2018selfbleu}. \\[0.4cm]

Information Entropy (IE) $\uparrow$ & \( E_{\text{Entro}} = - \sum_{i=1}^{n} p_i \log(p_i) \) & The measure of the randomness of the distribution~\cite{berger-etal-1996-maximum}, where $p_i$ is the probability/frequency of a token. \\[0.01cm]
\bottomrule
\end{tabular}
}
\label{tab:metrics}
\end{table*}



\textbf{Model Training and Evaluations.}
Each series of datasets is then fine-tuned from the same pretrained model, where Llama-2-7B~\cite{2023llama2} is used for most of our fine-tuning experiments.

For the evaluations, we adopt the pairwise scoring methodology used in Arena Hard~\cite{li2024crowdsourced} and combine the testing samples of both AlpacaEval 2.0 and Arena Hard benchmarks~\cite{alpaca_eval} for a more comprehensive evaluation.
GPT-4 Turbo is originally used by the Arena Hard scoring system.
However, due to its high cost for our extensive evaluations, we replace the judge by Llama-3.1-70B-Nemotron~\cite{wang2024Nemotron}, which shows a very high degree of agreement, with a reversal rate of only 8{\%}.
Additional evaluations using alternative judge models further confirm this observation. 
Further details regarding the evaluation setup and the consistency analysis are provided in the Appendix~\ref{subsec:judgemodel}. 


At the end of this step, we obtain the scores $\mathcal{S}$ with respect to constructed datasets, \textit{i.e.}, $\{\mathcal{D}_m, \mathcal{S}_m\}_{m=k_1, k_2, \cdots, k_M}$.
Results on these two items across multiple diversity scales ($M$), three dataset sizes (10K, 20K and 30K), three levels(macro-, meso- and microscopic) and two components (instructions and responses), as well as detailed analyses and ablation studies, will be introduced in the following two Sections.

\textbf{Posterior Diversity Metrics.}
Previous works have proposed to measure the diversity by the metrics listed in Table~\ref{tab:metrics} that provide quantitative insights into different aspects in posterior, while some of them are used or can be optimized when constructing datasets.
Moreover, we would like to mention the metric of the information entropy~\cite{berger-etal-1996-maximum} that may be a reasonable metric.
Detailed discussions are provided in Section~\ref{subsec:discussions}.

\vspace{-0.1cm}
\section{Main Results}
\vspace{-0.1cm}
In this section, we conduct the comparative experiments as introduced in the previous section, to compare the tuples of datasets and the corresponding scores.
To briefly recall the framework, there are three aspects that experiments will be conducted for the comparison:

\textbf{(1) Across multiple diversity scales.} 
For a certain size of datasets (e.g., 10K), we vary the diversity scale from the lowest to the highest that the strategy can achieve.
This range of diversity defines the bounds, where we manually set the lowest as 0\% and the highest as 100\%.
The values in-between represent different diversity scales.
This definition 
makes it possible to compare across strategies.

\textbf{(2) Across three dataset sizes.} 
We choose three sizes of constructed datasets, \textit{i.e.}, 10K, 20K and 30K from a corpus of 117K instructions and refreshed responses.

\textbf{(3) Across three strategic levels and two components.} 
We apply macro-, meso- and microscopic dataset diversity strategies on instructions and responses, to perform a comprehensive comparison.


At the end of this section, we furthermore test the effectiveness of the listed posterior diversity metrics for measuring the correlation between diversity and performance.





\vspace{-0.1cm}
\subsection{Comparative Results}

\begin{figure*}[t]
\centering 
\includegraphics[width=0.95\textwidth]{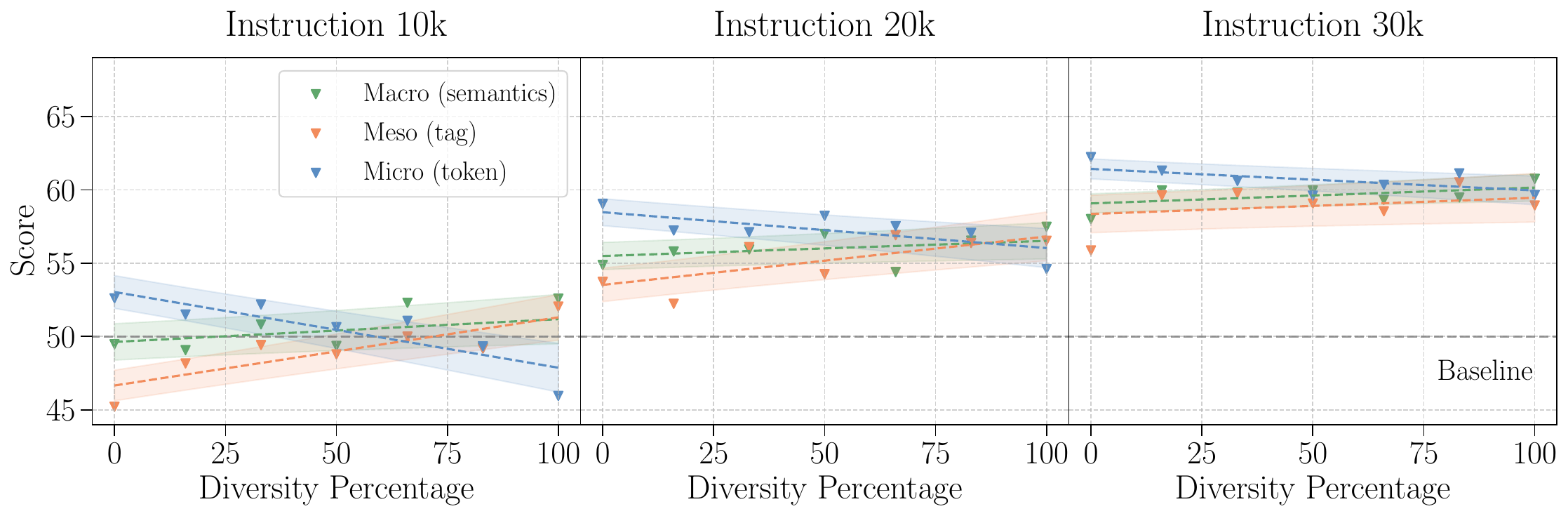}
\includegraphics[width=0.95\textwidth]{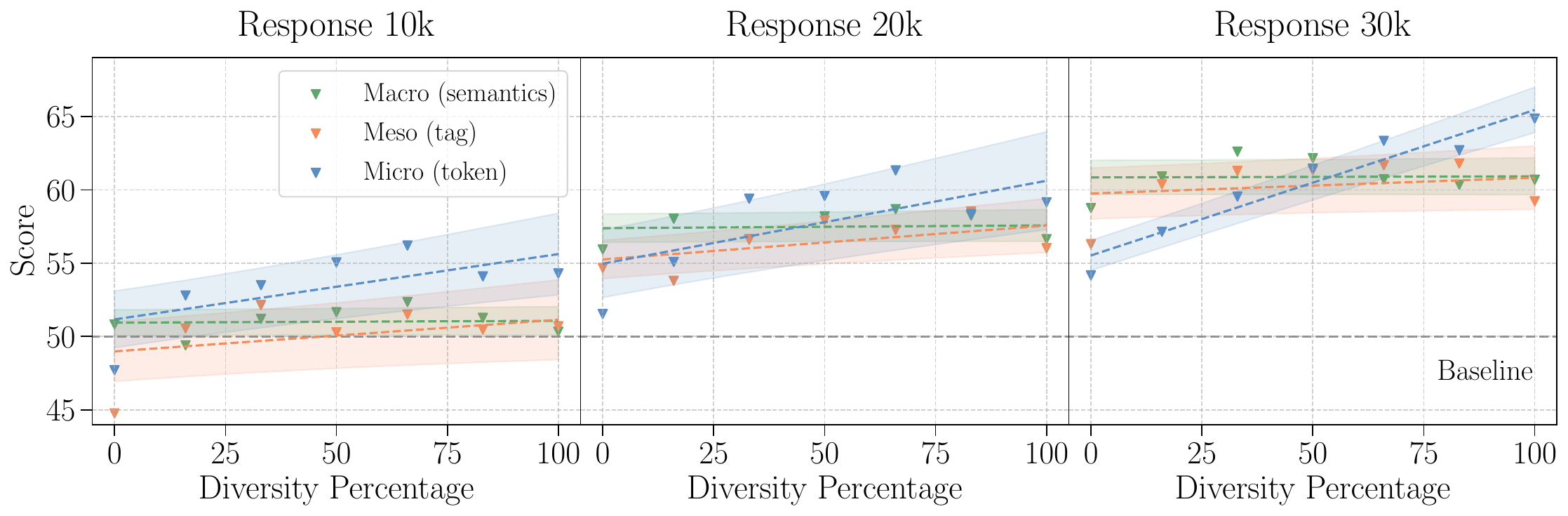}
\vspace{-0.2cm}
\caption{The relationship between the diversity percentage and model performance for instructions and responses across three dataset sizes (10K, 20K, and 30K). The vertical axis shows the average score from two benchmarks, with a baseline score of 50, derived based on a randomly selected 10K dataset. In each subplot, different colors correspond to three diversity levels: Macro (semantics), Meso (tag), and Micro (token). The markers represent the data, while the dashed lines and shaded areas indicate the results of Bayesian linear regression with $1$-$\sigma$ uncertainty.}
\label{fig:Instruction_response_variety}
\vspace{-0.2cm} 
\end{figure*}

We show the main results in Figure~\ref{fig:Instruction_response_variety}, where the observations and findings are discussed below.
Note that as the pairwise scoring system requires a reference baseline, a random 10K dataset is used for all experiments in this section.
An equal model scores 50 according to this scoring system, so dashed gray lines are drawn at the score of 50.

\vspace{-0.1cm}
\subsubsection{Dataset Sizes}
Our experiments with constrained dataset sizes (10K, 20K and 30K training samples) reveal a consistent performance trend where models trained on larger datasets generally outperform those using smaller training datasets.
While we hypothesize that diminishing returns or performance plateaus might emerge at greater dataset magnitudes, systematic investigation of this phenomenon remains beyond the scope of the current study.

Nevertheless, increased dataset diversity helps mitigate the performance gap associated with varying dataset sizes. 
Notably, some smaller yet more diverse datasets curated on the response component at the token level achieve superior model performance compared to larger yet less diverse datasets.
This finding underscores the importance and effectiveness of dataset diversity during dataset construction.


\vspace{-0.3cm}
\subsubsection{Macro-to-Micro Diversity Strategies}

Combined with the statistic results in Table~\ref{tab:slope} which quantify the slopes of lines in the semantic diversities and scores of fine-tuned LLM models.
This aligns with the hypothesis that instruction sets encompassing broader semantic Figure~\ref{fig:Instruction_response_variety}, we observe that the \textbf{macroscopic} strategy (semantics level) on instructions show positive correlations between ranges enhance a model's generalization capability.
A similar yet weaker correlation is observed for the response component.
However, compared to another two strategies, the range of performance scores that the macroscopic strategy can achieve is narrow, suggesting that the macroscopic strategy is less effective than finer-grained strategies.
\begin{wraptable}{r}{0.5\textwidth}
\centering
\vspace{-0.3cm}
\caption{Slope ($\times10^{-2}$) of the linear regression fit in Figure~\ref{fig:Instruction_response_variety} for each diversity-control strategy.}
\vspace{0.2cm}
\label{tab:slope}
\resizebox{0.5\columnwidth}{!}{%
\begin{tabular}{@{}ccccccc@{}}
\toprule
Dataset & \multicolumn{3}{c}{Instruction} & \multicolumn{3}{c}{Response} \\
\cmidrule(lr){2-4} \cmidrule(l){5-7}
Size & Macro & Meso & Micro & Macro & Meso & Micro \\
\midrule
10K & 2.37 & \underline{4.93} & $-$ 5.45 & 0.73 & 3.61 & \textbf{5.35} \\
20K & 1.67 & \underline{3.75} & $-$ 2.83 & 0.97 & 1.71 & \textbf{6.68} \\
30K & 1.45 & \underline{2.07} & $-$ 1.80 & 0.59 & 2.00 & \textbf{10.06} \\
\bottomrule
\end{tabular}
}
\vspace{-0.3cm}
\end{wraptable}

\vspace{-0.2cm}
Results of the \textbf{mesoscopic} strategy (tag level) show consistently positive correlation between the diversity scales and scores, on both instructions and responses.
The mesoscopic strategy shows larger slopes than the macroscopic, demonstrating that decomposing the entire instruction into functional attributes of the text, such as topics, intents, sentiments, areas \textit{etc} is more effective.


The proposed \textbf{microscopic} strategy (token-level) demonstrates limited efficacy when applied to instructions but exhibits the most significant impact on responses compared to the previous two strategies. This differential performance aligns with expectations. 
The effectiveness on responses stems from the fact that LLMs are explicitly supervised on response tokens during the SFT stage. 
Since LLMs generate outputs autoregressively (token-by-token), introducing token diversity likely mitigates overfitting by forcing the model to generalize across varied tokens, making the training more robust and generalizable.

Conversely, the strategy’s failure on instructions is intuitively explainable. 
First, token-level diversity in instructions does not inherently translate to meaningful semantic diversity. 
A microscopic focus on tokens risks overlooking broader contextual and semantic relationships. 
Second, we hypothesize that successful LLM alignment depends on consistent exposure to recurring token patterns in instructions. 
Reducing the frequency of specific tokens (via diversification) may dilute critical alignment signals, weakening the model’s ability to internalize task-specific linguistic or behavioral norms.

\vspace{-0.1cm}
\subsection{Tests of Diversity Metrics}
\label{subsec:metrics}

\begin{wrapfigure}{r}{0.55\textwidth}
\centering 
\vspace{-0.4cm}
\includegraphics[width=0.5\columnwidth]{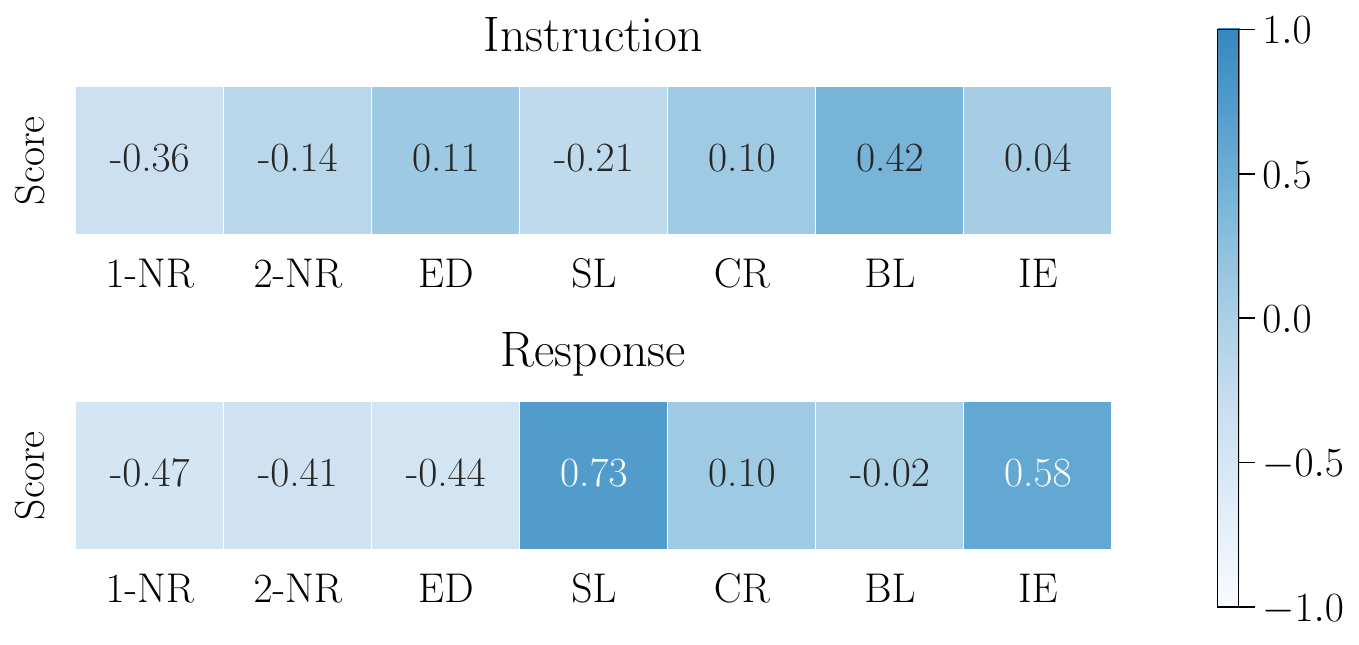}
\vspace{-0.25cm}
\caption{Pearson correlation coefficients of multiple diversity parameters and model performance scores based on all the involved 10K datasets. The upper and lower columns represent the diversity measurements from the perspectives of instruction and response, respectively. The intersection of two parameters shows their Pearson correlation coefficients.}
\label{fig:relation_10K}
\vspace{-0.35cm} 
\end{wrapfigure}

Most of the investigated metrics, as detailed in Section~\ref{subsec:framework} and Table~\ref{tab:metrics}, cannot be easily optimized during the dataset construction, and thus require to be tested in a posterior way.
For experiments per dataset size, we compute the metric values and measure the correlation to performance scores.
Figure~\ref{fig:relation_10K} shows the results.
Note that the metrics are applied to the instruction and the response respectively.

For the instruction, the most correlated metric is Self-BLEU (0.42).
This coheres with the macroscopic strategy on instructions.
Both would like to reserve diverse instructions, while the macroscopic applies the semantically clustering way and Self-BLEU measures the semantic similarities.
However, as noticed previously, the macroscopic strategy is less effective than meso- and microscopic strategies.

Turning to response-level metrics, sequence length (0.73) and information entropy (0.58) exhibit the strongest correlations with performance scores. 
The high correlation of sequence length aligns with prior findings that prioritizing longer responses serves as an effective heuristic for enhancing model performance in SFT datasets~\cite{zhao2024longlength}. 
To investigate whether this correlation reflects direct causation, we conduct ablation experiments in Section~\ref{subsec:length_control} using the proposed microscopic strategy on responses filtering with strict length constraints. 
By maintaining near-identical lengths across compared datasets, the ablation study isolates length as a variable to assess its true impact on model outcomes.

The information entropy metric relates to the core mechanism of our microscopic strategy, with the slight difference that the strategy optimizes the entropy over the set of mid-band tokens and the metric computes over the whole vocabulary set.
Moreover, the strong correlation with information entropy implies the effectiveness of the microscopic strategy.
The entropy measures the token distribution.
When it is computed on responses, it measures the distribution of training signals.
High entropy means diverse tokens and can help the LLM to be less prone to the overfitting, making the training more robust and generalizable. Detailed entropy analyses are provided in Appendix~\ref{subsec:entropy}.


\vspace{-0.1cm}
\section{Ablation Studies}
\label{sec: Ablation Studies}

Main experimental results indicate that macroscopic and mesoscopic strategies effectively lead to higher performance with increasing diversity, whereas the microscopic strategy on response exhibits both a stronger correlation between model performance and diversity, and superior performance with maximum diversity across all strategies. 
To further investigate the mechanisms of the proposed strategy, we conduct three ablation studies in this section.


\vspace{-0.1cm}
\subsection{Model}
Building on our observations with Llama-2-7B, we conduct a cross-model analysis to examine whether the performance advantages of microscopic response diversity generalize across different model scales and architectures. We select two contrasting models for this investigation: the larger Llama-2-70B and the more recent Llama-3-8B.
For each architecture, we construct two series of datasets of 10K samples through the microscopic strategy on response using their respective tokenizers, and train the models on these datasets.
To enable meaningful comparison, we establish two baselines with either model respectively using randomly sampled datasets of equivalent size (10K samples) without diversity optimization and conduct the comparison.
\begin{wrapfigure}{r}{0.5\textwidth}
\centering 
\includegraphics[width = 0.5\columnwidth]{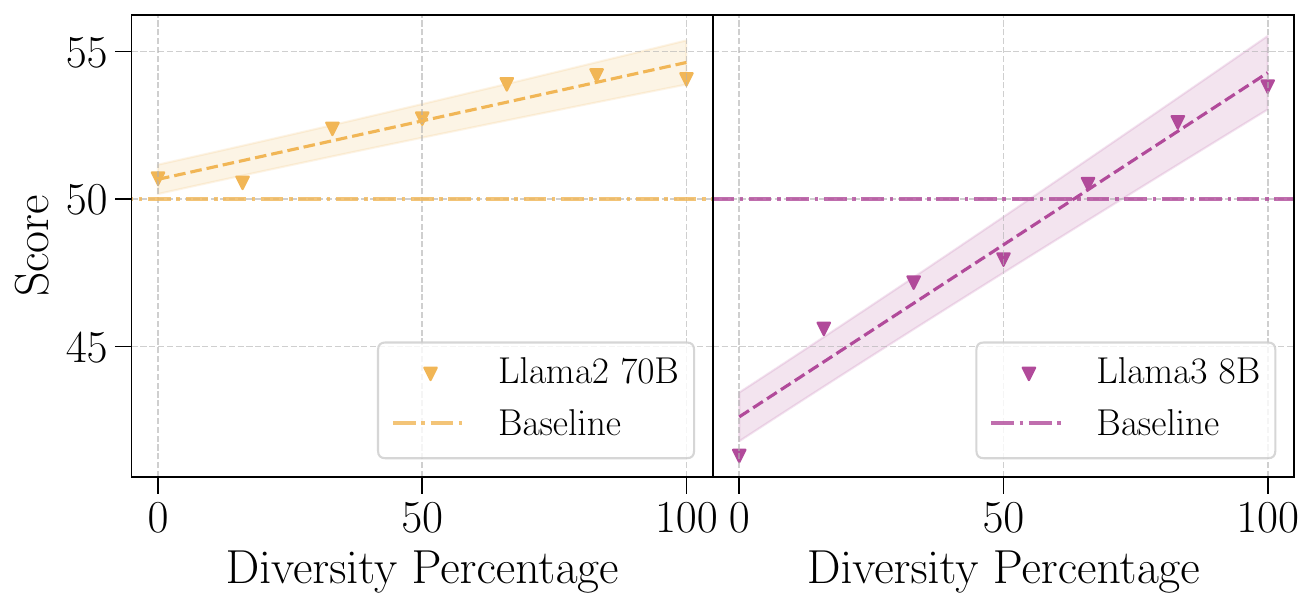}
\vspace{-0.4cm}
\caption{Ablation study on microscopic diversity: comparing different models (Llama-2-70B and Llama-3-8B). The abscissa indicates the microscopic level diversity percentage from the response perspective in the case of 10K dataset.}
\label{fig:abl_model}
\vspace{-0.2cm} 
\end{wrapfigure}

As evidenced in Figure~\ref{fig:abl_model}, both Llama-2-70B and Llama-3-8B mirror the performance pattern of Llama-2-7B. 
More strikingly, Llama 3-8B demonstrates an amplified correlation slope, suggesting heightened sensitivity to token-level diversity variations. 
While future works are required to investigate the deep reasons, we posit this difference stems from Llama 3's implementation of the tiktoken tokenizer, which enlarges the vocabulary size from Llama-2's 32K to 128K and the covering range of mid band tokens.
Overall, these results confirm a robust and consistent positive correlation across different model sizes and architectures.


\vspace{-0.2cm}
\subsection{Length Control}
\label{subsec:length_control}
As discussed in Section~\ref{subsec:metrics}, response length exhibits a strong correlation with model performance scores.
To investigate whether this relationship is causal, we conducted experiments controlling the response length to be identical.
This ablation study uses the microscopic strategy on responses.
Specially, we construct length-controlled  datasets by sampling from sub-datasets of varying lengths and selecting samples within a similar length range.
Models are subsequently fine-tuned on these datasets and evaluated across diversity scales as presented in the proposed framework.

Figure~\ref{fig:ablation_length} (a) reveals that while we restricted response lengths to approximately 500 tokens across all datasets, the relationship between diversity scale and model performance remains unchanged compared to the uncontrolled baseline. 
This persistent correlation suggests that response length may serve as an incidental covariate rather than the primary factor driving performance improvements.

\vspace{-0.1cm}
\subsection{Tokenizer}

\begin{wrapfigure}{r}{0.6\textwidth}
	\centering
        \vspace{-0.5cm}
	\subfigure[Length Control] {\includegraphics[width=.29\textwidth]{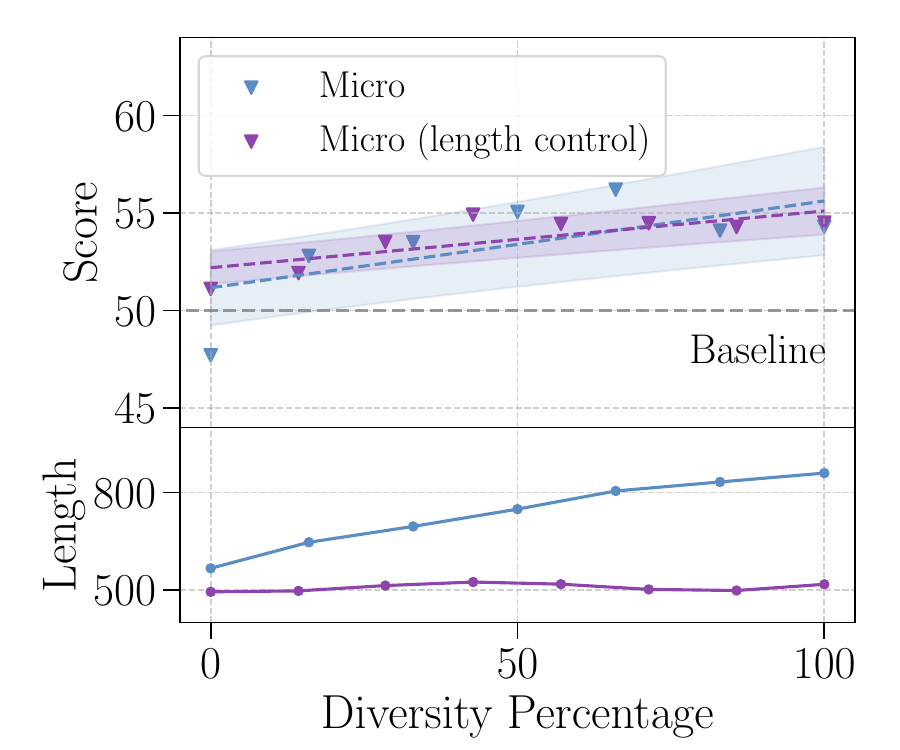}}
	\subfigure[Tokenizer Comparison] {\includegraphics[width=.29\textwidth]{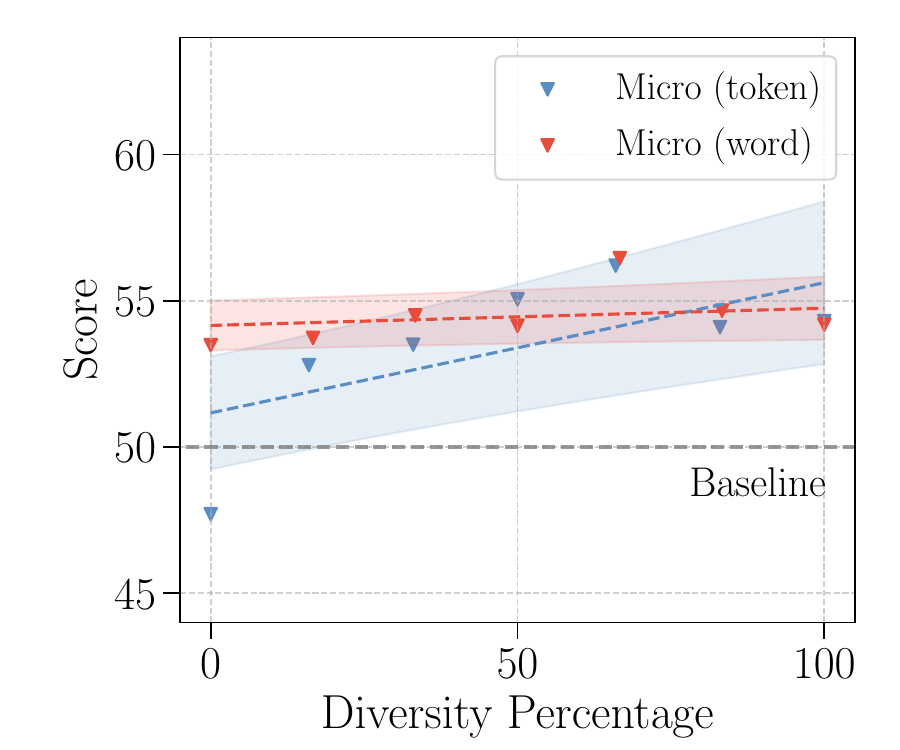}}
    \vspace{-0.2cm}
    \caption{Ablation study on microscopic diversity: Figure (a) compares the performance under length control. The subplot below illustrates the average token length of responses across the datasets. Figure (b) compares word-based tokenization. The abscissa indicates the microscopic level diversity percentage from the response perspective in the 10K datasets.}
    \label{fig:ablation_length}
    \vspace{-0.4cm}
\end{wrapfigure}


Given the crucial role of tokenization in the microscopic strategy, we investigate its influence by comparing the default tokenization scheme with word segmentation. 
The default tokenization aligns with the LLM to be fine-tuned while the word segmentation is a common approach that can be used for any LLM without any pre-processing step.
We thus only replace the tokenize step by the word segmentation in the microscopic strategy, and construct a new series of datasets (of 10K samples) to train the model.



As shown in Figure~\ref{fig:ablation_length} (b), word-based tokenization does not exhibit a clear positive correlation with model performance.
This suggests that the tokenization scheme provided by the tokenizer is essential for preserving microscopic diversity, as it segments information in a way that matches input the model received during training. Word-based segmentation may disrupt this correspondence, leading to suboptimal performance. 
\vspace{-0.1cm}
\subsection{Additional Ablation Studies and Discussions}
\label{subsec:discussions}

Three additional studies are provided in Appendix~\ref{sec:additional_as} to further confirm our findings in various settings.
(1) With original responses (which are of lower data quality), (2) On more benchmarks (MMLU~\cite{mmlu} and LiveBench~\cite{livebench}) and (3) with mismatched tokenizer to control the microscopic dataset diversity.




While prior research has predominantly emphasized the instruction diversity that implicitly leads to diverse responses and consequently diverse token distributions, our comprehensive experiments reveal a more impactful approach.
Through rigorous comparative analysis, diversity metric evaluations, and ablation studies, we conclude that the microscopic strategy on response is a more effective and direct way.
We therefore encourage the research community to prioritize the response diversification alongside the instruction for SFT dataset construction.



\vspace{-0.1cm}
\section{Related Work}

\subsection{SFT Diversity Measure Approach}
Dataset diversity plays a crucial role in the SFT stage for LLM development. Research in this area can be broadly categorized into two main approaches:

\textbf{Quantitative Feature Approach.}
Numerous studies explored fundamental dataset features to evaluate their correlation with language model performance.
Commonly examined features include distinct-n~\cite{li-etal-2016-distinctn}, quantifying unique n-grams proportion as a dataset diversity measure; 
gzip compression ratio~\cite{song2024scaling}, indicating data redundancy; 
ROUGE-L score~\cite{lin-2004-rouge}, measuring sequence overlap; 
and Self-BLEU~\cite{zhu2018selfbleu}, assessing the internal diversity of text samples. 
While these features provide intuitive quantitative metrics for dataset diversity, they are often applied in relatively simple ways, limiting their capacity to capture deeper contextual relationships within the data.

\textbf{Semantic Representation Approach.}
Advances in language models~\cite{2017attention} have made analyzing dataset semantic structures a critical focus, prompting methods to encode textual information into high-dimensional representations.
Bidirectional Encoder Representations from Transformers (BERT)~\cite{devlin2019bert} model has been widely adopted for this purpose. 
For instance, embeddings derived from BERT have been leveraged to assess dataset diversity by quantifying various semantic properties, including the distance of embedding vectors~\cite{shaib2024standardizing, liu2024DEITA, wang2024diversity, liu2024tsds} and the identification of semantically coherent clusters of instructions using clustering techniques~\cite{grootendorst2022bertopic,ge-etal-2024-car}. 
These approaches offer valuable insights into the semantic diversity present within datasets.

The rapid development of generative autoregressive models~\cite{zhang-etal-2020-dialogpt, brown2020gpt3, achiam2023gpt} has shifted research towards leveraging these models for both semantic and diversity analysis. 
LLMs are now increasingly employed to extract richer semantic features, such as generating descriptive tags for instructional data~\cite{lu2023instag, yang2024qwen2, dubey2024llama3} and identifying clustering patterns through LLM-based agents~\cite{chen2024onthescalediversity}. 
This offers a complementary approach to traditional methods for assessing dataset diversity by leveraging the nuanced semantic understanding provided by these models.

\vspace{-0.2cm}
\subsection{Token-Level Analysis in Large Language Models}
Token-level analysis has been studied in LLMs, with numerous approaches exploring the relationship between tokens and various aspects of LLM training and performance~\cite{madsen2022evaluatingtokens,  zhong2023revisiting, li2024entropicmatch, land2024fishing, lin2024rho}. 
For instance, Lin et al. \cite{lin2024rho} focus on training strategies that prioritize important tokens to enhance model efficiency. Madsen et al. \cite{madsen2022evaluatingtokens} have observed variations in training time and gradient descent loss associated with processing different tokens, using these metrics to infer token importance or influence. Li et al. \cite{li2024entropicmatch} have explored the use of token matching techniques to improve training efficiency and performance.
While these studies provide valuable insights into token-level effects on training, the analysis of token diversity within the context of SFT datasets remains relatively unexplored.

\vspace{-0.2cm}
\section{Conclusion}
In this study, we first propose a taxonomy to categorize the approaches for dataset diversity according to the grain level (macro-, meso- and microscopic) and the component of SFT datasets (instructions and responses).
Through extensive experiments, we systematically compare their impact on model performance and demonstrate that our proposed microscopic strategy on responses exhibits the strongest correlation between model performance and diversity degree while achieving superior performance at maximum diversity compared to other strategies.
Detailed analyses and ablation studies confirm the effectiveness of the proposed microscopic strategy on responses.
We have also tested multiple diversity metrics and suggest that the information entropy may be a good estimator of dataset diversity, which also aligns with the optimization direction of the microscopic strategy.
We therefore call upon the research community to prioritize the response diversification alongside the instruction for SFT dataset construction.

\bibliographystyle{abbrv}
{\footnotesize
\bibliography{ref.bib}

\begin{thebibliography}{10}

\bibitem{abusamra2013gini}
H.~Abusamra.
\newblock A comparative study of feature selection and classification methods for gene expression data of glioma.
\newblock {\em Procedia Computer Science}, 23:5--14, 2013.

\bibitem{achiam2023gpt}
J.~Achiam, S.~Adler, S.~Agarwal, L.~Ahmad, I.~Akkaya, F.~L. Aleman, D.~Almeida, J.~Altenschmidt, S.~Altman, S.~Anadkat, et~al.
\newblock Gpt-4 technical report.
\newblock {\em arXiv preprint arXiv:2303.08774}, 2023.

\bibitem{arora2017embeddingdis}
S.~Arora, Y.~Liang, and T.~Ma.
\newblock A simple but tough-to-beat baseline for sentence embeddings.
\newblock In {\em Proceedings of the International Conference on Learning Representations}, 2017.

\bibitem{berger-etal-1996-maximum}
A.~L. Berger, S.~A. Della~Pietra, and V.~J. Della~Pietra.
\newblock A maximum entropy approach to natural language processing.
\newblock {\em Computational Linguistics}, 22(1):39--71, 1996.

\bibitem{brown2020gpt3}
T.~Brown, B.~Mann, N.~Ryder, M.~Subbiah, J.~D. Kaplan, P.~Dhariwal, A.~Neelakantan, P.~Shyam, G.~Sastry, A.~Askell, et~al.
\newblock Language models are few-shot learners.
\newblock {\em Advances in Neural Information Processing Systems}, 33:1877--1901, 2020.

\bibitem{bubeck2023sparks}
S.~Bubeck, V.~Chandrasekaran, R.~Eldan, J.~Gehrke, E.~Horvitz, E.~Kamar, P.~Lee, Y.~T. Lee, Y.~Li, S.~Lundberg, et~al.
\newblock Sparks of artificial general intelligence: Early experiments with gpt-4.
\newblock {\em arXiv preprint arXiv:2303.12712}, 2023.

\bibitem{bukharin2023QDIT}
A.~Bukharin, S.~Li, Z.~Wang, J.~Yang, B.~Yin, X.~Li, C.~Zhang, T.~Zhao, and H.~Jiang.
\newblock Data diversity matters for robust instruction tuning.
\newblock In {\em Findings of the 2024 Conference on Empirical Methods in Natural Language Processing}, pages 3411--3425, 2024.

\bibitem{chen2024onthescalediversity}
H.~Chen, A.~Waheed, X.~Li, Y.~Wang, J.~Wang, B.~Raj, and M.~I. Abdin.
\newblock On the diversity of synthetic data and its impact on training large language models.
\newblock {\em arXiv preprint arXiv:2410.15226}, 2024.

\bibitem{DatabricksBlog2023DollyV2}
M.~Conover, M.~Hayes, A.~Mathur, J.~Xie, J.~Wan, S.~Shah, A.~Ghodsi, P.~Wendell, M.~Zaharia, and R.~Xin.
\newblock Free dolly: Introducing the world's first truly open instruction-tuned llm, 2023.

\bibitem{dettmers2024qlora}
T.~Dettmers, A.~Pagnoni, A.~Holtzman, and L.~Zettlemoyer.
\newblock Qlora: Efficient finetuning of quantized llms.
\newblock {\em Advances in Neural Information Processing Systems}, 36, 2023.

\bibitem{devlin2019bert}
J.~Devlin, M.-W. Chang, K.~Lee, and K.~Toutanova.
\newblock Bert: Pre-training of deep bidirectional transformers for language understanding.
\newblock In {\em Proceedings of the 2019 Conference of the North American Chapter of the Association for Computational Linguistics: Human Language Technologies}, pages 4171--4186, 2019.

\bibitem{du2023mods}
Q.~Du, C.~Zong, and J.~Zhang.
\newblock Mods: Model-oriented data selection for instruction tuning.
\newblock {\em arXiv preprint arXiv:2311.15653}, 2023.

\bibitem{dubey2024llama3}
A.~Dubey, A.~Jauhri, A.~Pandey, A.~Kadian, A.~Al-Dahle, A.~Letman, A.~Mathur, A.~Schelten, A.~Yang, A.~Fan, et~al.
\newblock The llama 3 herd of models.
\newblock {\em arXiv preprint arXiv:2407.21783}, 2024.

\bibitem{ester1996dbscan}
M.~Ester, H.-P. Kriegel, J.~Sander, X.~Xu, et~al.
\newblock A density-based algorithm for discovering clusters in large spatial databases with noise.
\newblock In {\em Proceedings of the 2nd International Conference on Knowledge Discovery and Data Mining}, volume~96, pages 226--231, 1996.

\bibitem{ge2024scaling}
T.~Ge, X.~Chan, X.~Wang, D.~Yu, H.~Mi, and D.~Yu.
\newblock Scaling synthetic data creation with 1,000,000,000 personas.
\newblock {\em arXiv preprint arXiv:2406.20094}, 2024.

\bibitem{ge-etal-2024-car}
Y.~Ge, Y.~Liu, C.~Hu, W.~Meng, S.~Tao, X.~Zhao, M.~Xia, Z.~Li, B.~Chen, H.~Yang, B.~Li, T.~Xiao, and J.~Zhu.
\newblock Clustering and ranking: Diversity-preserved instruction selection through expert-aligned quality estimation.
\newblock In {\em Proceedings of the 2024 Conference on Empirical Methods in Natural Language Processing}, pages 464--478, 2024.

\bibitem{grootendorst2022bertopic}
M.~Grootendorst.
\newblock Bertopic: Neural topic modeling with a class-based tf-idf procedure.
\newblock {\em arXiv preprint arXiv:2203.05794}, 2022.

\bibitem{mmlu}
D.~Hendrycks, C.~Burns, S.~Basart, A.~Zou, M.~Mazeika, D.~Song, and J.~Steinhardt.
\newblock Measuring massive multitask language understanding.
\newblock In {\em Proceedings of the International Conference on Learning Representations}, 2021.

\bibitem{hu2022lora}
E.~J. Hu, yelong shen, P.~Wallis, Z.~Allen-Zhu, Y.~Li, S.~Wang, L.~Wang, and W.~Chen.
\newblock Lo{RA}: Low-rank adaptation of large language models.
\newblock In {\em Proceedings of the International Conference on Learning Representations}, 2022.

\bibitem{kaplan2020scaling}
J.~Kaplan, S.~McCandlish, T.~Henighan, T.~B. Brown, B.~Chess, R.~Child, S.~Gray, A.~Radford, J.~Wu, and D.~Amodei.
\newblock Scaling laws for neural language models.
\newblock {\em arXiv preprint arXiv:2001.08361}, 2020.

\bibitem{land2024fishing}
S.~Land and M.~Bartolo.
\newblock Fishing for magikarp: Automatically detecting under-trained tokens in large language models.
\newblock In {\em Proceedings of the 2024 Conference on Empirical Methods in Natural Language Processing}, pages 11631--11646, 2024.

\bibitem{li-etal-2016-distinctn}
J.~Li, M.~Galley, C.~Brockett, J.~Gao, and B.~Dolan.
\newblock A diversity-promoting objective function for neural conversation models.
\newblock In {\em Proceedings of the 2016 Conference of the North American Chapter of the Association for Computational Linguistics: Human Language Technologies}, pages 110--119, 2016.

\bibitem{li2024crowdsourced}
T.~Li, W.-L. Chiang, E.~Frick, L.~Dunlap, T.~Wu, B.~Zhu, J.~E. Gonzalez, and I.~Stoica.
\newblock From crowdsourced data to high-quality benchmarks: Arena-hard and benchbuilder pipeline.
\newblock {\em arXiv preprint arXiv:2406.11939}, 2024.

\bibitem{alpaca_eval}
X.~Li, T.~Zhang, Y.~Dubois, R.~Taori, I.~Gulrajani, C.~Guestrin, P.~Liang, and T.~B. Hashimoto.
\newblock Alpacaeval: An automatic evaluator of instruction-following models.
\newblock \url{https://github.com/tatsu-lab/alpaca_eval}, 2023.

\bibitem{li2024entropicmatch}
Z.~Li, C.~Chen, T.~Xu, Z.~Qin, J.~Xiao, R.~Sun, and Z.-Q. Luo.
\newblock Entropic distribution matching for supervised fine-tuning of {LLM}s: Less overfitting and better diversity.
\newblock In {\em Proceedings of the 38th Annual Conference on Neural Information Processing Systems Workshop on Fine-Tuning in Modern Machine Learning: Principles and Scalability}, 2024.

\bibitem{lin-2004-rouge}
C.-Y. Lin.
\newblock {ROUGE}: A package for automatic evaluation of summaries.
\newblock In {\em Proceedings of Text Summarization Branches Out}, pages 74--81, July 2004.

\bibitem{lin2024rho}
Z.~Lin, Z.~Gou, Y.~Gong, X.~Liu, yelong shen, R.~Xu, C.~Lin, Y.~Yang, J.~Jiao, N.~Duan, and W.~Chen.
\newblock Not all tokens are what you need for pretraining.
\newblock In {\em Proceedings of the thirty-eighth Annual Conference on Neural Information Processing Systems}, 2024.

\bibitem{liu2024DEITA}
W.~Liu, W.~Zeng, K.~He, Y.~Jiang, and J.~He.
\newblock What makes good data for alignment? a comprehensive study of automatic data selection in instruction tuning.
\newblock In {\em Proceedings of the International Conference on Learning Representations}, 2024.

\bibitem{liu2024tsds}
Z.~Liu, A.~Karbasi, and T.~Rekatsinas.
\newblock {TSDS}: Data selection for task-specific model finetuning.
\newblock In {\em Proceedings of the Annual Conference on Neural Information Processing Systems}, 2024.

\bibitem{lu2023instag}
K.~Lu, H.~Yuan, Z.~Yuan, R.~Lin, J.~Lin, C.~Tan, C.~Zhou, and J.~Zhou.
\newblock \# instag: Instruction tagging for analyzing supervised fine-tuning of large language models.
\newblock In {\em Proceedings of the International Conference on Learning Representations}, 2023.

\bibitem{madsen2022evaluatingtokens}
A.~Madsen, N.~Meade, V.~Adlakha, and S.~Reddy.
\newblock Evaluating the faithfulness of importance measures in nlp by recursively masking allegedly important tokens and retraining.
\newblock In {\em Findings of the 2022 Conference on Empirical Methods in Natural Language Processing}, pages 1731--1751, 2022.

\bibitem{mcinnes2017hdbscan}
L.~McInnes, J.~Healy, and S.~Astels.
\newblock hdbscan: Hierarchical density based clustering.
\newblock {\em Journal of Open Source Software}, 2(11):205, 2017.

\bibitem{mcinnes2018umap}
L.~McInnes, J.~Healy, N.~Saul, and L.~Gro{\ss}berger.
\newblock Umap: Uniform manifold approximation and projection.
\newblock {\em Journal of Open Source Software}, 3(29):861, 2018.

\bibitem{mikolov2013word2vec}
T.~Mikolov, I.~Sutskever, K.~Chen, G.~S. Corrado, and J.~Dean.
\newblock Distributed representations of words and phrases and their compositionality.
\newblock {\em Advances in Neural Information Processing Systems}, 26, 2013.

\bibitem{ouyang2022training}
L.~Ouyang, J.~Wu, X.~Jiang, D.~Almeida, C.~Wainwright, P.~Mishkin, C.~Zhang, S.~Agarwal, K.~Slama, A.~Ray, et~al.
\newblock Training language models to follow instructions with human feedback.
\newblock {\em Advances in Neural Information Processing Systems}, 35, 2022.

\bibitem{padmakumar2024does_ngramdiv}
V.~Padmakumar and H.~He.
\newblock Does writing with language models reduce content diversity?
\newblock In {\em Proceedings of the International Conference on Learning Representations}, 2024.

\bibitem{shaib2024standardizing}
C.~Shaib, J.~Barrow, J.~Sun, A.~F. Siu, B.~C. Wallace, and A.~Nenkova.
\newblock Standardizing the measurement of text diversity: A tool and a comparative analysis of scores.
\newblock {\em arXiv preprint arXiv:2403.00553}, 2024.

\bibitem{song2024scaling}
F.~Song, B.~Yu, H.~Lang, H.~Yu, F.~Huang, H.~Wang, and Y.~Li.
\newblock Scaling data diversity for fine-tuning language models in human alignment.
\newblock In {\em Proceedings of the 2024 Joint International Conference on Computational Linguistics, Language Resources and Evaluation}, pages 14358--14369, 2024.

\bibitem{2023llama2}
H.~Touvron, L.~Martin, K.~Stone, P.~Albert, A.~Almahairi, Y.~Babaei, N.~Bashlykov, S.~Batra, P.~Bhargava, S.~Bhosale, et~al.
\newblock Llama 2: Open foundation and fine-tuned chat models.
\newblock {\em arXiv preprint arXiv:2307.09288}, 2023.

\bibitem{2017attention}
A.~Vaswani, N.~Shazeer, N.~Parmar, J.~Uszkoreit, L.~Jones, A.~N. Gomez, L.~u. Kaiser, and I.~Polosukhin.
\newblock Attention is all you need.
\newblock {\em Advances in Neural Information Processing Systems}, 30, 2017.

\bibitem{wang2024diversity}
P.~Wang, Y.~Shen, Z.~Guo, M.~Stallone, Y.~Kim, P.~Golland, and R.~Panda.
\newblock Diversity measurement and subset selection for instruction tuning datasets.
\newblock {\em arXiv preprint arXiv:2402.02318}, 2024.

\bibitem{wang2023self}
Y.~Wang, Y.~Kordi, S.~Mishra, A.~Liu, N.~A. Smith, D.~Khashabi, and H.~Hajishirzi.
\newblock Self-instruct: Aligning language models with self-generated instructions.
\newblock In {\em Proceedings of the 61st Annual Meeting of the Association for Computational Linguistics}, pages 13484--13508, 2023.

\bibitem{wang2024Nemotron}
Z.~Wang, A.~Bukharin, O.~Delalleau, D.~Egert, G.~Shen, J.~Zeng, O.~Kuchaiev, and Y.~Dong.
\newblock Helpsteer2-preference: Complementing ratings with preferences.
\newblock {\em arXiv preprint arXiv:2410.01257}, 2024.

\bibitem{wei2022emergent}
J.~Wei, Y.~Tay, R.~Bommasani, C.~Raffel, B.~Zoph, S.~Borgeaud, D.~Yogatama, M.~Bosma, D.~Zhou, D.~Metzler, et~al.
\newblock Emergent abilities of large language models.
\newblock {\em Transactions on Machine Learning Research}, 2022.

\bibitem{livebench}
C.~White, S.~Dooley, M.~Roberts, A.~Pal, B.~Feuer, S.~Jain, R.~Shwartz-Ziv, N.~Jain, K.~Saifullah, S.~Dey, Shubh-Agrawal, S.~S. Sandha, S.~V. Naidu, C.~Hegde, Y.~LeCun, T.~Goldstein, W.~Neiswanger, and M.~Goldblum.
\newblock Livebench: A challenging, contamination-free {LLM} benchmark.
\newblock In {\em Proceedings of the International Conference on Learning Representations}, 2025.

\bibitem{wiggins2022alpaca}
W.~F. Wiggins and A.~S. Tejani.
\newblock On the opportunities and risks of foundation models for natural language processing in radiology.
\newblock {\em Radiology: Artificial Intelligence}, 4(4):e220119, 2022.

\bibitem{yang2024qwen2}
A.~Yang, B.~Yang, B.~Zhang, B.~Hui, B.~Zheng, B.~Yu, C.~Li, D.~Liu, F.~Huang, H.~Wei, et~al.
\newblock Qwen2. 5 technical report.
\newblock {\em arXiv preprint arXiv:2412.15115}, 2024.

\bibitem{zhang-etal-2020-dialogpt}
Y.~Zhang, S.~Sun, M.~Galley, Y.-C. Chen, C.~Brockett, X.~Gao, J.~Gao, J.~Liu, and B.~Dolan.
\newblock {DIALOGPT} : Large-scale generative pre-training for conversational response generation.
\newblock In {\em Proceedings of the 58th Annual Meeting of the Association for Computational Linguistics: System Demonstrations}, July 2020.

\bibitem{pmlr-v235-zhao24a}
D.~Zhao, J.~Andrews, O.~Papakyriakopoulos, and A.~Xiang.
\newblock Position: Measure dataset diversity, don’t just claim it.
\newblock In {\em Proceedings of the 41st International Conference on Machine Learning}, pages 60644--60673, 2024.

\bibitem{zhao2024longlength}
H.~Zhao, M.~Andriushchenko, F.~Croce, and N.~Flammarion.
\newblock Long is more for alignment: A simple but tough-to-beat baseline for instruction fine-tuning.
\newblock In {\em Proceedings of the 41st International Conference on Machine Learning}, 2024.

\bibitem{zheng2024llamafactory}
Y.~Zheng, R.~Zhang, J.~Zhang, Y.~Ye, Z.~Luo, Z.~Feng, and Y.~Ma.
\newblock Llamafactory: Unified efficient fine-tuning of 100+ language models.
\newblock In {\em Proceedings of the 62nd Annual Meeting of the Association for Computational Linguistics: System Demonstrations}, 2024.

\bibitem{zhong2023revisiting}
Q.~Zhong, L.~Ding, J.~Liu, X.~Liu, M.~Zhang, B.~Du, and D.~Tao.
\newblock Revisiting token dropping strategy in efficient bert pretraining.
\newblock In {\em Proceedings of the 61st Annual Meeting of the Association for Computational Linguistics}, pages 10391--10405, 2023.

\bibitem{zhou2024lima}
C.~Zhou, P.~Liu, P.~Xu, S.~Iyer, J.~Sun, Y.~Mao, X.~Ma, A.~Efrat, P.~Yu, L.~Yu, et~al.
\newblock Lima: Less is more for alignment.
\newblock {\em Advances in Neural Information Processing Systems}, 36, 2023.

\bibitem{zhu2018selfbleu}
Y.~Zhu, S.~Lu, L.~Zheng, J.~Guo, W.~Zhang, J.~Wang, and Y.~Yu.
\newblock Texygen: A benchmarking platform for text generation models.
\newblock In {\em Proceedings of the 41st International ACM SIGIR Conference on Research \& Development in Information Retrieval}, pages 1097--1100, 2018.

\end{thebibliography}
}

\appendix
\onecolumn
\section{Collection of Open-Source SFT Datasets}
\label{subsec:dataset}
For transparency and reproducibility, we disclose the publicly available SFT datasets used in our work, including:

\textbf{Alpaca}~\cite{wiggins2022alpaca}: The Alpaca dataset consists of 52K instruction-response samples generated by text-davinci-003 using the Self-Instruct method~\cite{wang2023self}, which is used to fine-tune Llama.

\textbf{ShareGPT}\footnote{The original ShareGPT dataset (\url{https://sharegpt.com/}) has not been officially released.  Therefore, we utilize a reproduced version available on the Hugging Face (\url{https://huggingface.co/datasets/anon8231489123/ShareGPT_Vicuna_unfiltered}), applying a cleaning preprocessing step.}: The ShareGPT dataset consists of conversations scraped from the ShareGPT platform before its discontinuation. These conversations contain both user prompts and corresponding responses from ChatGPT, offering a diverse range of conversational contexts.

\textbf{Dolly}~\cite{DatabricksBlog2023DollyV2}: The Dolly dataset, released by Databricks as part of the Dolly project, is an open-source dataset of human-generated instruction-following examples, licensed for research and commercial use, and used to fine-tune Dolly2.0 models.

\textbf{LIMA}~\cite{zhou2024lima}: The LIMA dataset contains 1,000 carefully curated instructions from diverse domains, focusing on high-quality prompts designed to elicit specific behaviors from language models.

Note that we reserve only the instructions from these datasets and generate the responses using Llama-3.1-70B-Nemotron model~\cite{wang2024Nemotron}.

\vspace{-0.1cm}
\section{Consistency Analysis of Judge Models}
\label{subsec:judgemodel}
We evaluate the judge agreement between Nemotron and GPT-4 Turbo through a comparative analysis of 500 pairwise comparisons from the Arena Hard benchmark. To minimize positional bias, the judge model evaluates each pair of responses twice, reversing the order of presentation in the second evaluation. Pairwise scores are calculated according to the criteria outlined in Table~\ref{tab:consistency_rule}. These two scores are summed and then categorized into one of three categories: Good, Same, or Bad. Subsequently, we compare the judgment consistency of the models against that of GPT-4 Turbo. The results reveal that the Llama-3.1-70B-Nemotron model~\cite{wang2024Nemotron} demonstrates strong agreement with GPT-4 Turbo, with a reversal rate of only 8\%. The pairwise evaluation prompt is shown in Appendix~\ref{app:judge_prompt}.

\vspace{-0.2cm}
\begin{table}[h!]
\centering
\caption{Model judgment score compute rules.}
\begin{tabular}{c ccccc}
\toprule
Score & 1 & 0.5 & 0 & -0.5 & -1 \\
\midrule
Judge1    & A\textgreater\textgreater B & A\textgreater B & A=B & B\textgreater A & B\textgreater\textgreater A \\
Judge2    & B\textgreater\textgreater A & B\textgreater A & A=B & A\textgreater B & A\textgreater\textgreater B \\
\bottomrule
\end{tabular}
\label{tab:consistency_rule}
\end{table}

In spite of this, we conduct additional experiments to assess the robustness of our conclusions and mitigate concerns regarding model-specific biases. Specifically, we evaluate the 10K micro-level response set using two structurally distinct judge models: Qwen2.5-72B-Instruct and GPT-4o. As shown in Table~\ref{tab:abl_judgemodel}, the evaluation scores obtained from these alternative judges exhibit consistent trends, reinforcing that the observed effects are not dependent on a particular evaluation model.

\vspace{-0.1cm}
\begin{table}[h!]

\centering
\caption{Evaluation scores under different judge models across micro-level diversity.}

\begin{tabular}{l cccc}
\toprule

Micro Diversity (\%) &	0  &33.3	&66.7 &100\\

\midrule
GPT-4o	&48.4	&51.8	&52.8	&53.1\\
Qwen2.5-72B-Instruct	&49.0	&54.1	&55.4	&57.4\\

\bottomrule
\end{tabular}
\label{tab:abl_judgemodel}
\end{table}
\vspace{-0.3cm}

\section{Additional Ablation Studies}
\label{sec:additional_as}

In addition to the ablation analyses presented in Section~\ref{sec: Ablation Studies}, we include several supplementary experiments that are crucial for understanding the overall findings.
\subsection{Performance on Origin Response}

To verify the rationale for response replacement, we evaluated our strategy's effectiveness using the original responses. Specifically, we performed supplementary experiments by applying our micro-level diversity strategy to these baseline responses.The results in Table~\ref{tab:response_compare} consistently show that response diversity positively correlates with model performance when compared with the original 10K response baseline. We also find that the original responses group showed significantly lower scores in the nemotron 10K baseline in preliminary comparisons.

\vspace{-0.2cm}
\begin{table}[h!]
\centering

\caption{Performance with different levels of micro-level diversity using original responses. The baseline derived from random 10K original response samples.}
\begin{tabular}{l ccccccc}
\toprule

Micro Diversity (\%) &	0 &	16.7	&33.3	&50	&66.7	&83.3	&100\\

\midrule
Arena-Hard Score	&37.6	&43.8	&48.5	&50.8	&55.1	&59.4	&61.4\\
AlpacaEval 2.0 Score	&43.3	&49.1	&57.2	&61.5	&64.9	&66.4	&64.5\\

\bottomrule
\end{tabular}
\label{tab:response_compare}
\end{table}
\vspace{-0.3cm}

\subsection{Performance on Other Benchmarks}
\label{subsec:otherbenchmark}

Given that the evaluation in our study relies on LLM-based judgments, we further assess model performance using alternative benchmarks—MMLU and LiveBench—which employ objective, non-LLM-based evaluation metrics. We conduct this evaluation using the microscopic strategy on response 10K datasets with the Llama-3-8B model. As shown in Table~\ref{tab:otherbench}, the results significantly indicate that greater response diversity leads to improved model performance across both benchmarks.

\begin{table}[h!]
\centering

\caption{Performance on MMLU and LiveBench across different levels of micro-level diversity.}
\begin{tabular}{l ccccccc}
\toprule

Micro Diversity (\%) &	0 &	16.7	&33.3	&50	&66.7	&83.3	&100\\

\midrule
MMLU Score	&54.96	&52.62	&54.42	&52.27	&56.31	&56.50	&55.68\\
LiveBench Score	&12.14	&13.27	&13.13	&12.28	&14.55	&13.75	&14.47\\

\bottomrule
\end{tabular}
\label{tab:otherbench}
\end{table}
\vspace{-0.3cm}

\subsection{Robustness under Tokenizer Mismatch}
\label{subsec:tokenizer_mismatch}
To further assess the robustness of our method under variations in tokenization, we perform an additional experiment summarized in Table~\ref{tab:mismatch_tokenizer}. In this setting, diversity selection is conducted using the Llama-2 tokenizer, while model training is carried out with Llama-3-8B. Notably, these tokenizers differ significantly in both vocabulary and tokenization schemes. While the performance stability does decline under this mismatch—as expected—our method continues to yield notable improvements, indicating its resilience to tokenizer variation. Nevertheless, we acknowledge that optimal results are achieved when the tokenizer and model are well aligned.

\begin{table}[h!]
\centering

\caption{Model performance under tokenizer mismatch: Llama-2's tokenizer used for data strategy and Llama-3-8B for training.}
\begin{tabular}{l ccccccc}
\toprule

Micro Diversity (\%) &	0 &	16.7	&33.3	&50	&66.7	&83.3	&100\\

\midrule
Arena-Hard Score	&45.0	&51.1	&45.9	&52.6	&54.1	&56.9	&58.5\\
AlpacaEval 2.0 Score	&41.6	&50.7	&45.5	&58.1	&51.4	&61.6	&53.8\\
\bottomrule
\end{tabular}
\label{tab:mismatch_tokenizer}
\end{table}
\vspace{-0.3cm}

\section{Experimental Parameters and Setup}
\label{subsec:setup}
To facilitate reproducibility, we list the parameters and provide details for the three levels of diversity strategy applied to the datasets, as well as the training settings used in the SFT training process.

\vspace{-0.1cm}
\subsection{Macroscopic Parameters}
At the macroscopic scale, we utilize the all-MiniLM-L12-v2 model for embeddings. The UMAP parameters are configured with neighbors=15 and components=5, and HDBSCAN clustering is performed using the Euclidean distance metric with a minimum cluster size of 20. This process results in approximately 2,000 semantic clusters for both instructions and responses. To construct the dataset, we sample instruction-response pairs from a fixed number of clusters to ensure balanced representation across clusters. The macroscopic diversity is defined as the number of selected clusters.

\vspace{-0.2cm}

\subsection{Mesoscopic Parameters}
At the mesoscopic scale, we employ Llama-3.1-70B-Nemotron~\cite{wang2024Nemotron} for tagging, with prompt details provided in Appendix~\ref{app:tag_prompt}. We then perform a consistency check using Qwen2.5-72B-Instruct~\cite{yang2024qwen2}. Following this, we filter out non-words and merge synonymous terms. The resulting tags are clustered using HDBSCAN with eps=0.15 and min samples=2, yielding approximately 5,000 tag clusters for both instructions and responses. These tags are subsequently normalized by their respective cluster center tags. The mesoscopic diversity is defined as the ratio of distinct tag categories to the total number of tags.

\subsection{Microscopic Parameters}
At the microscopic scale, tokens are extracted using the tokenizer from the training model. Tokens that appear with frequencies between 10 and 500 are defined as ``important tokens'', as they are assumed to have a significant impact on the dataset distribution. To systematically investigate how token diversity affects model behavior, our experiment employs two sequential stages: The preparatory Algorithm~\ref{alg:stage0} constructs distinct diversity subsets through inverse greedy pruning, followed by the core sampling process in Algorithm~\ref{alg:stage1} that selects data based on the distribution patterns of these important tokens. Microscopic diversity is quantified as the proportion of unique important tokens relative to the total important token types.

\vspace{-0.2cm}

\begin{algorithm}[H]
\caption{Inverse Greedy Pruning}
\label{alg:stage0}
\begin{algorithmic}[1]
   \STATE \textbf{Input:} 
       Dataset $\mathcal{D} = \{d_1,...,d_n\}$ where each $d_i$ has token set $T_{d_i}$,
       Target token count $K$
   \STATE \textbf{Output:} 
       Pruned subset $\mathcal{D}_{\text{sub}}$,
       Temporary tokens $T_{\text{tmp}}$
       
   \STATE $\mathcal{D}_{\text{sub}} \gets \mathcal{D}$ \COMMENT{Initialize with full dataset}
   \STATE $T_{\text{tmp}} \gets \bigcup_{d \in \mathcal{D}} T_d$ \COMMENT{Initial token coverage}
   
   \WHILE{$|T_{\text{tmp}}| > K$ \textbf{and} $\mathcal{D}_{\text{sub}} \neq \emptyset$}
       
       \STATE Calculate $\text{Unique}(d) = \big| T_d \setminus \bigcup_{d' \neq d} T_{d'} \big|$, $\forall d \in \mathcal{D}_{\text{sub}}$

       \STATE Select $d^* \gets \arg\max \text{Unique}(d)$
       
       \STATE Remove sample: $\mathcal{D}_{\text{sub}} \gets \mathcal{D}_{\text{sub}} \setminus \{d^*\}$
       \STATE Update coverage: $T_{\text{tmp}} \gets \bigcup_{d \in \mathcal{D}_{\text{sub}}} T_d$
   \ENDWHILE
   
   \STATE \textbf{return} $\mathcal{D}_{\text{sub}}$, $T_{\text{tmp}}$
\end{algorithmic}
\end{algorithm}
\vspace{-0.4cm}

\begin{algorithm}[H]
\caption{Integrated Token-Aware Sampling}
\label{alg:stage1}
\begin{algorithmic}[1]
   \STATE \textbf{Input:} 
       Subset $\mathcal{D}_{\text{sub}}$,
       Candidate tokens $T_{\text{tmp}}$,
       Target size $N$,
       Trade-off $\alpha > 0$,
       Batch size $B$
   \STATE \textbf{Output:} 
       Final sample $\mathcal{S}$
       
   \STATE Initialize $\mathcal{S} \gets \emptyset$, $\mathcal{R} \gets \mathcal{D}_{\text{sub}}, \text{Counts}[t] \gets 0$, $\forall t \in T_{\text{tmp}}$
   
   \WHILE{$|\mathcal{S}| < N$ \textbf{AND} $\mathcal{R} \neq \emptyset$}
       \IF{$\exists t \in T_{\text{tmp}}$ with $\text{Counts}[t] = 0$}  
           \STATE Find $d^* \in \mathcal{R}$ maximizing new coverage:
           \STATE $d^* \gets \arg\max |T_d \cap \{t | \text{Counts}[t] = 0\}|$, $\forall d \in \mathcal{R}$
       \ELSE  
           \STATE Compute scores: $s(d) \gets \sum_{t \in T_d}\frac{1}{\text{Counts}[t]+\alpha}$, $\forall d \in \mathcal{R}$
           \STATE Sort $\mathcal{R}$ descending by $s(d)$
           \STATE Batch Size $k \gets \min(B, N - |\mathcal{S}|)$
           \STATE Select top-$k$ documents $d^*_1,...,d^*_k$ from sorted $\mathcal{R}$
       \ENDIF
       
       \FOR{each $d^*$}  
           \STATE $\mathcal{S} \gets \mathcal{S} \cup \{d^*\}$
           \STATE $\mathcal{R} \gets \mathcal{R} \setminus \{d^*\}$
           
           \STATE $\text{Counts}[t] \gets \text{Counts}[t] + 1$, $t \in T_{d^*}$
           
       \ENDFOR
   \ENDWHILE
   
   \STATE \textbf{return} $\mathcal{S}$
\end{algorithmic}
\end{algorithm}

\subsection{Training Setting}
\label{subsec:trainset}

We conduct full-parameter fine-tuning via Llama-Factory~\cite{zheng2024llamafactory} on 16×NVIDIA A800 80GB GPUs. To optimize computational efficiency, sequence packing is enabled and training sequences are then truncated to a maximum length of 4,096 tokens. 

For the learning rate schedule, we implement a cosine decay strategy starting at \(5 \times 10^{-6}\), with a 5\% warmup ratio to ensure stable initial training and avoid abrupt parameter updates. The model is fine-tuned over 4 epochs with a batch size of 16, striking a balance between computational efficiency and performance. Training a 7B model on 10K datasets requires approximately 20 minutes under this configuration.

\section{Correlation Analysis of Token Distribution Statistics and Model Performance}
\label{subsec:entropy}
\begin{figure*}[h!]
\centering 
\includegraphics[width=0.8\textwidth]{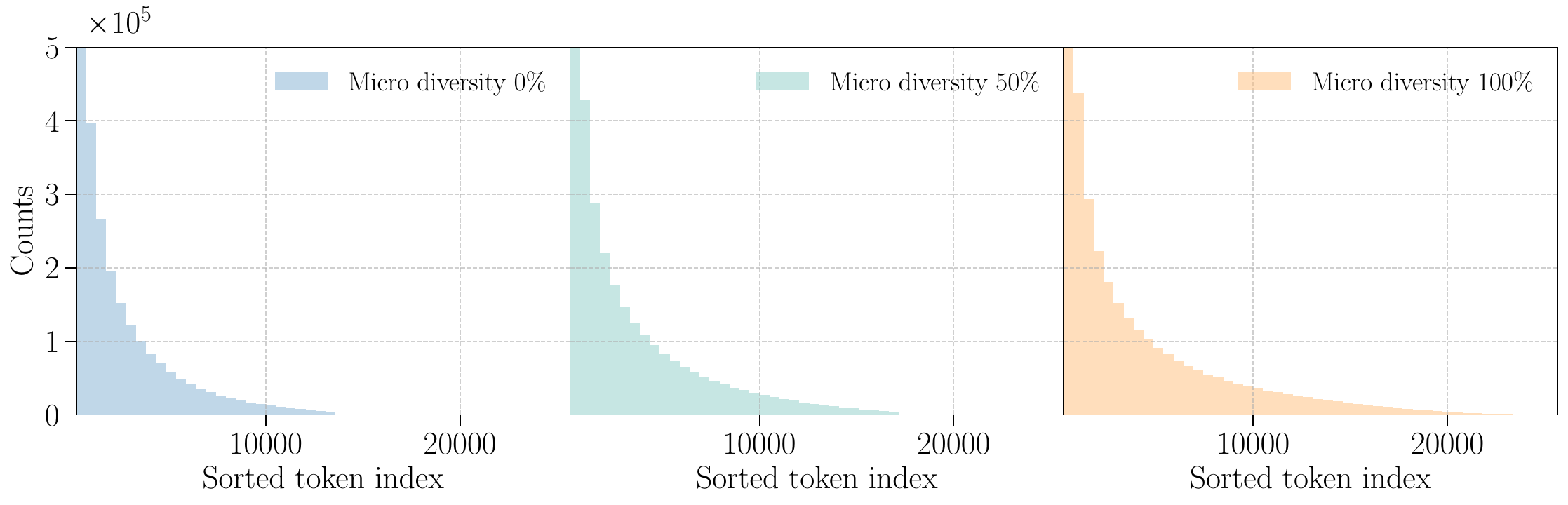}
\vspace{-0.4cm} 
\caption{Token distributions from the response perspective under different microscopic diversity percentages in the case of 10k dataset.}
\label{fig:appendix_token_distribution} 
\end{figure*}

Diversity-control strategies at the microscopic level actually alter the distribution of tokens. In the section, we explore the correlation between multiple statistics related to token distributions and model performance. Figure~\ref{fig:appendix_token_distribution} demonstrates the token distribution with different microscopic diversity percentage (0\%, 50\%, and 100\%) in the case of 10k dataset. Based on the distribution, we can explore various statistics related to token distributions to identify micro-level diversity parameters that might have a stronger correlation with model performance. We investigate the correlation between the IE (information entropy)~\cite{berger-etal-1996-maximum}, kurtosis, and Gini index~\cite{abusamra2013gini} based on token distributions and model performance from both the instruction and response perspectives. Information entropy quantifies how spread out the probabilities of different outcomes are, making it a statistical metric for evaluating diversity,
\begin{equation}
    {\rm Information\ entropy} = - \sum_{i=1}^{n} p_i \log(p_i)\,.
\end{equation}
Kurtosis describes the shape of a distribution, specifically its tailedness and the sharpness of its peak. It reads,
\begin{equation}
    {\rm Kurtosis} = \left[\frac{1}{n} \sum_{i=1}^{n} \left( \frac{x_i - \mu}{\sigma} \right)^{4}\right] - 3\,.
\end{equation}
The Gini index measures the inequality in a distribution. It reflects how evenly the probabilities are distributed among the tokens and is often used to evaluate imbalances,
\begin{equation}
    {\rm Gini\ index} = 1 - \sum_{i=1}^{n} p_i^{2}\,,
\end{equation}
where $p_i$ is the probability of $i$-th token, $x_i$ is the sorted index of $i$-th token. $\mu$ and $\sigma$ are the mean value and the standard deviation of the sorted indexes of tokens. The Pearson correlation coefficients between performance scores and these statistic are shown in Table~\ref{tab:distribution_diversity_scores}. Among these, IE from response perspective demonstrates the highest correlation. For the three statistics, the model performance score shows a positive correlation with IE, while exhibiting negative correlations with kurtosis and Gini index. All three metrics consistently indicate that flatter token distributions are beneficial for SFT of language models, confirming the feasibility of measuring diversity from the token level.

\vspace{-0.3cm}
\begin{table*}[h!]
\centering
\caption{Pearson correlation coefficients between the model performance score and various statistics based on token distributions.}
\vspace{0.1cm}
\begin{tabular}{c ccc ccc}
\toprule
Dataset & \multicolumn{3}{c}{Instruction} & \multicolumn{3}{c}{Response} \\
\cmidrule(r){2-4} \cmidrule(r){5-7}
Size & IE & $-$ Kurtosis & $-$ Gini index 
      & IE & $-$ Kurtosis & $-$ Gini index \\
\midrule
10K   & 0.04 & 0.29 & 0.01 & 0.58 & 0.58 & 0.42 \\
20K   & 0.13 & 0.52 & 0.11 & 0.62 & 0.43 & 0.50 \\
30K   & 0.25 & 0.68 & 0.28 & 0.76 & 0.32 & 0.69 \\
\bottomrule
\end{tabular}
\label{tab:distribution_diversity_scores}
\end{table*}

\vspace{1cm}

\vspace{-0.2cm} 

\section{Limitation and Future Works}

\label{sec:limitation}

Our study has provided initial insights into the role of dataset diversity in Language Model performance during Supervised Fine-Tuning (SFT), but several limitations and open questions remain. The scope of our experiments was constrained by the scale of available datasets and computational resources, limiting our analysis to SFT applications. Additionally, the relationship between dataset diversity and model performance, particularly in terms of entropy, requires more interpretable and comprehensive analysis to fully understand the underlying mechanisms.

\textbf{Mixture Strategy of Instruction and Response Diversity.}
A strategy that combines instruction and response diversity may yield more comprehensive outcomes. Our preliminary analysis of three strategies revealed minimal correlation between instruction-response dynamics, suggesting that a mixture strategy could enhance performance and clarify diversity patterns. This is also a direction we are considering for future exploration.

\textbf{Entropy and Dataset Diversity.}
Exploring the relationship between entropy and dataset diversity may lead to more comprehensive insights. While our findings suggest a potential link between language model performance and the entropy of SFT datasets, a more robust and interpretable analysis is needed to solidify these observations and provide deeper insights.

\textbf{Generalization to Alternative Fine-Tuning Paradigms}
While our current validation focuses on supervised learning frameworks, the possible relevance of our diversity-aware methodology to Reinforcement Learning environments warrants further investigation. Preliminary observations suggest that similar diversity control mechanisms could potentially be adapted for RL data sampling strategies, though this hypothesis requires systematic verification through dedicated studies.





\section{Raw Data of the Experiments}
In this section, we present the experimental data for the two benchmarks shown in Figure~\ref{fig:Instruction_response_variety}, as well as the detailed correlation analysis data for Figure~\ref{fig:relation_10K}.
\vspace{-0.2cm}
\subsection{Scores of each Benchmark}
\begin{figure*}[h!]
\centering 

\includegraphics[width=0.98\textwidth]{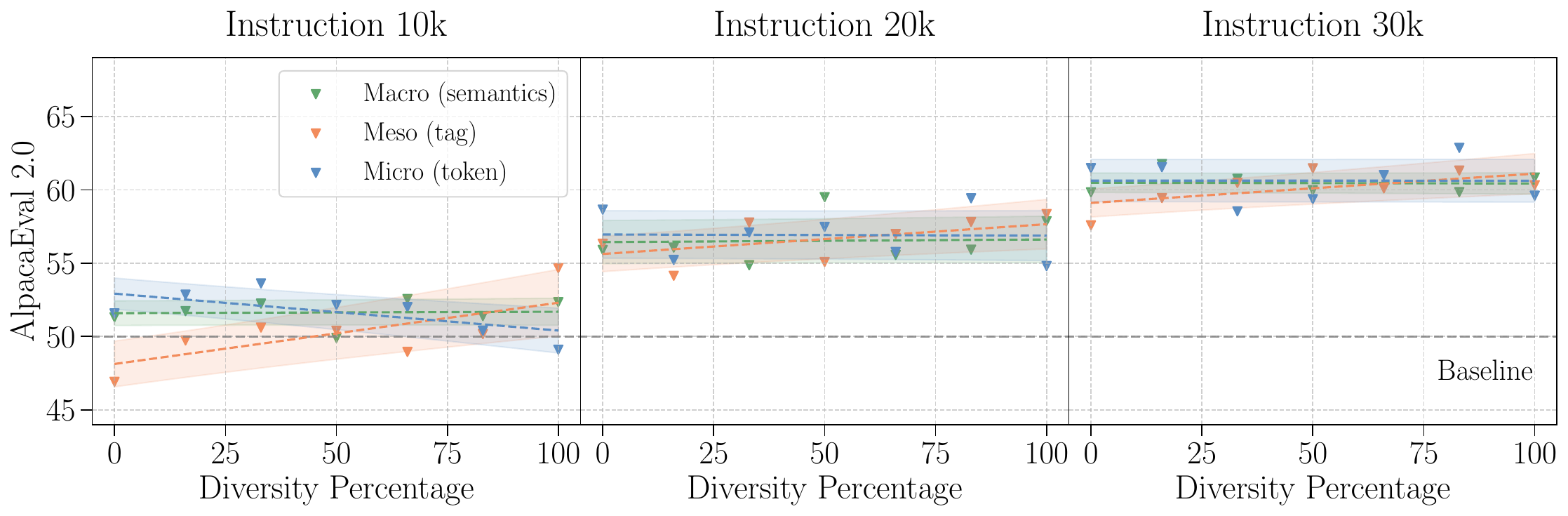}
\includegraphics[width=0.98\textwidth]{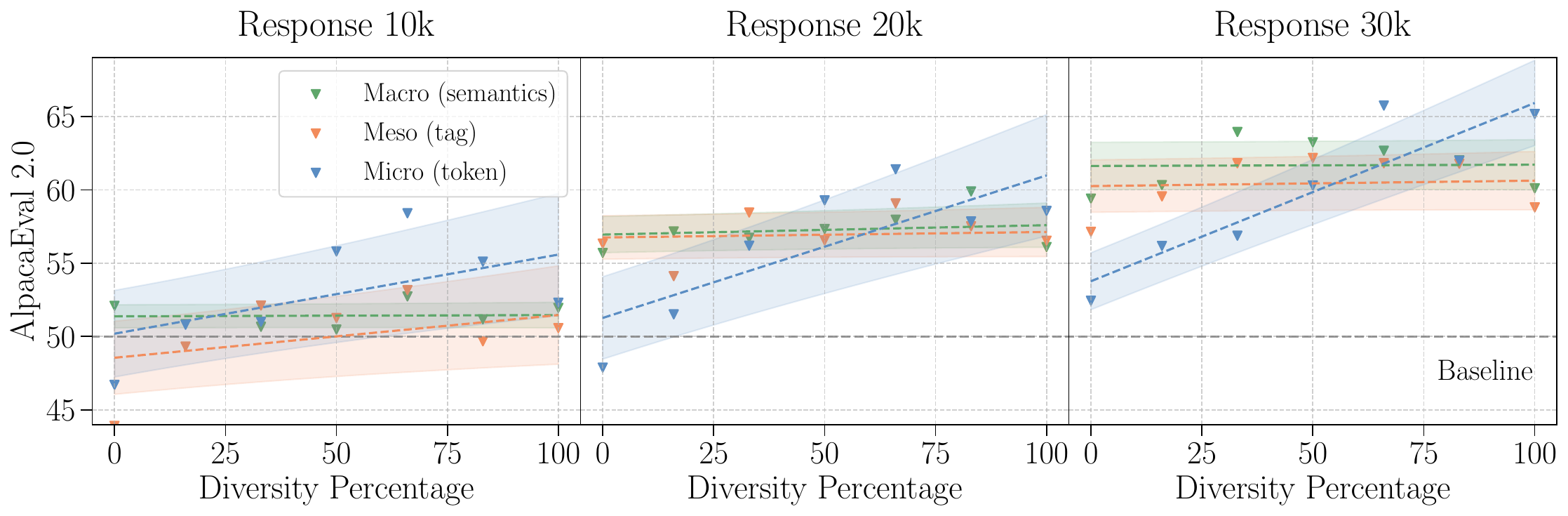}
\includegraphics[width=0.98\textwidth]{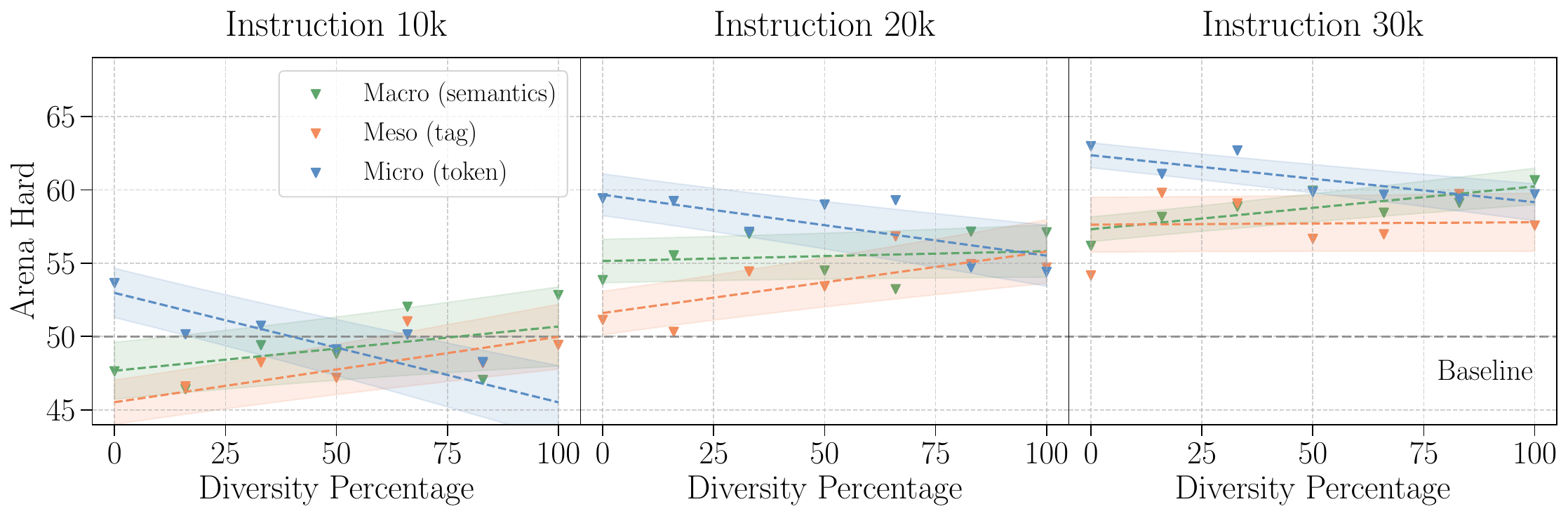}
\includegraphics[width=0.98\textwidth]{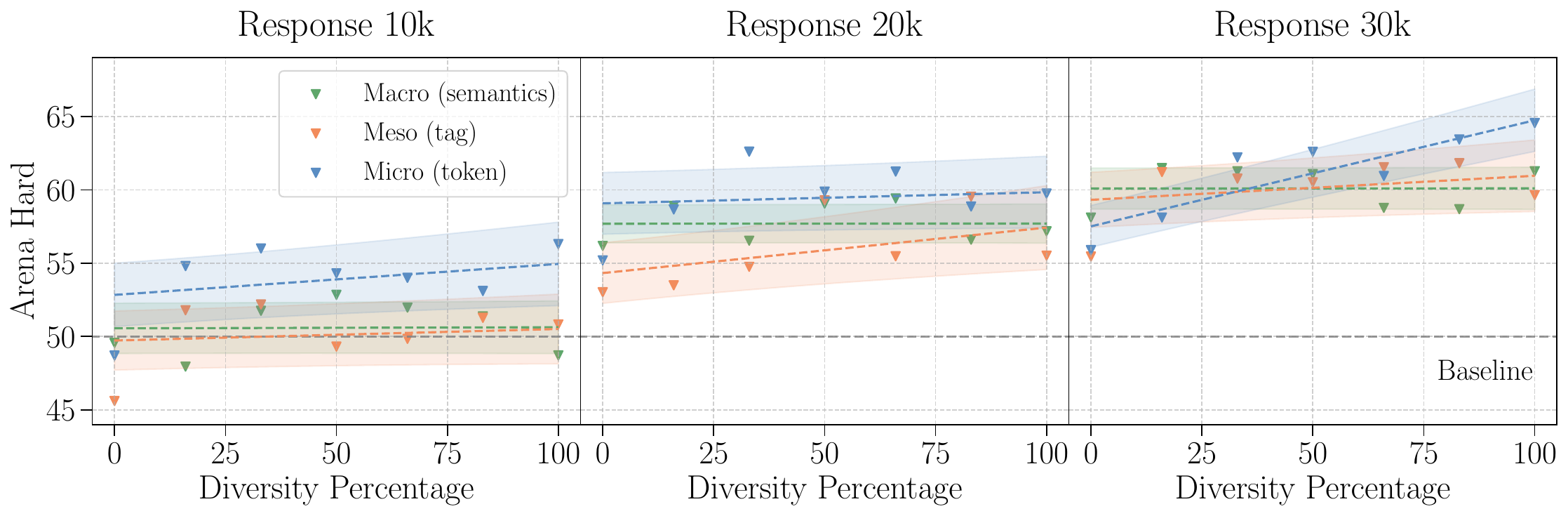}

\caption{The relationship between the diversity percentage and model performance for instructions and responses across three dataset sizes (10K, 20K, and 30K) in Arena Hard and AlpacaEval 2.0 benchmark. The settings and styles are the same as those in Figure~\ref{fig:Instruction_response_variety}.}
\label{fig:Instruction_response_variety_origin}
 
\end{figure*}

In Figure~\ref{fig:Instruction_response_variety_origin}, we present the experimental results separately for Arena Hard and AlpacaEval 2.0, providing a clear comparison of the scores across these two benchmarks.

\clearpage
\subsection{Correlation Analysis in Each Diversity-control Strategy}
Chapter~\ref{subsec:metrics} discusses the correlation between multiple diversity metrics and model performance across all 10K datasets. Here, we present the correlation figures for each diversity strategy, further categorized by instructions and responses, based on their corresponding 10K datasets.

\begin{figure*}[h!]
\centering 
\includegraphics[width=0.48\columnwidth]{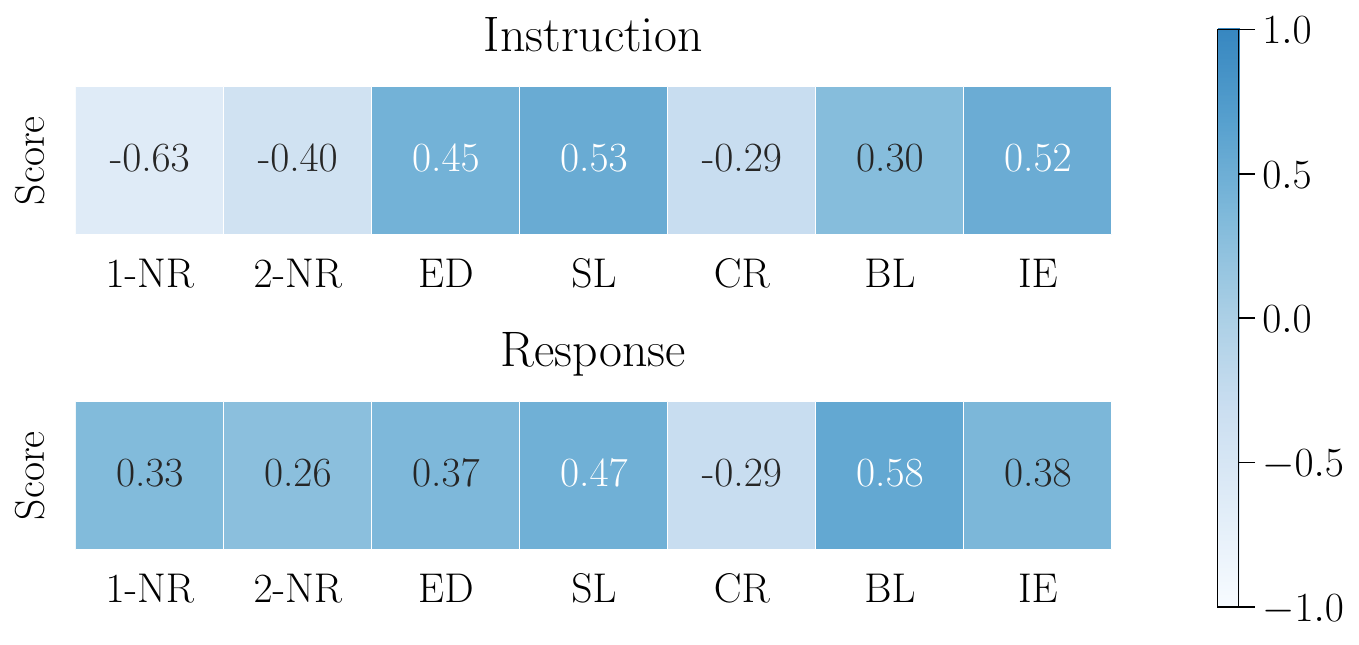}
\includegraphics[width=0.48\columnwidth]{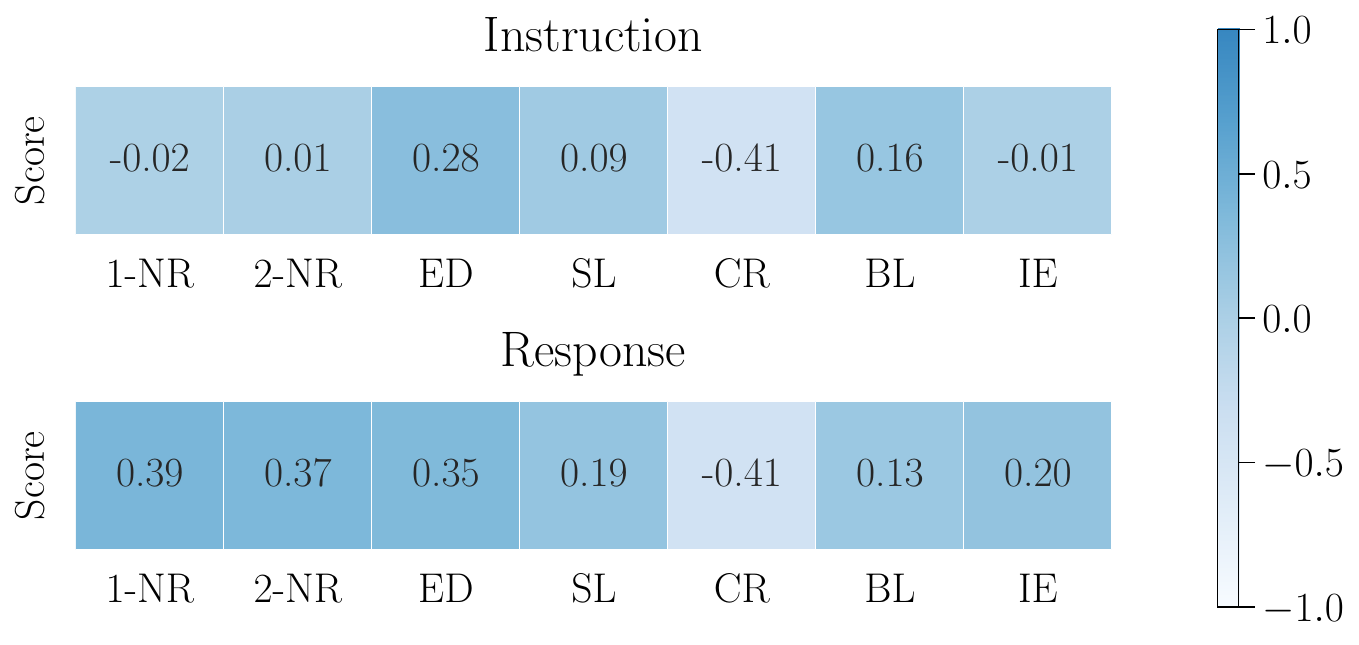}
\includegraphics[width=0.48\columnwidth]{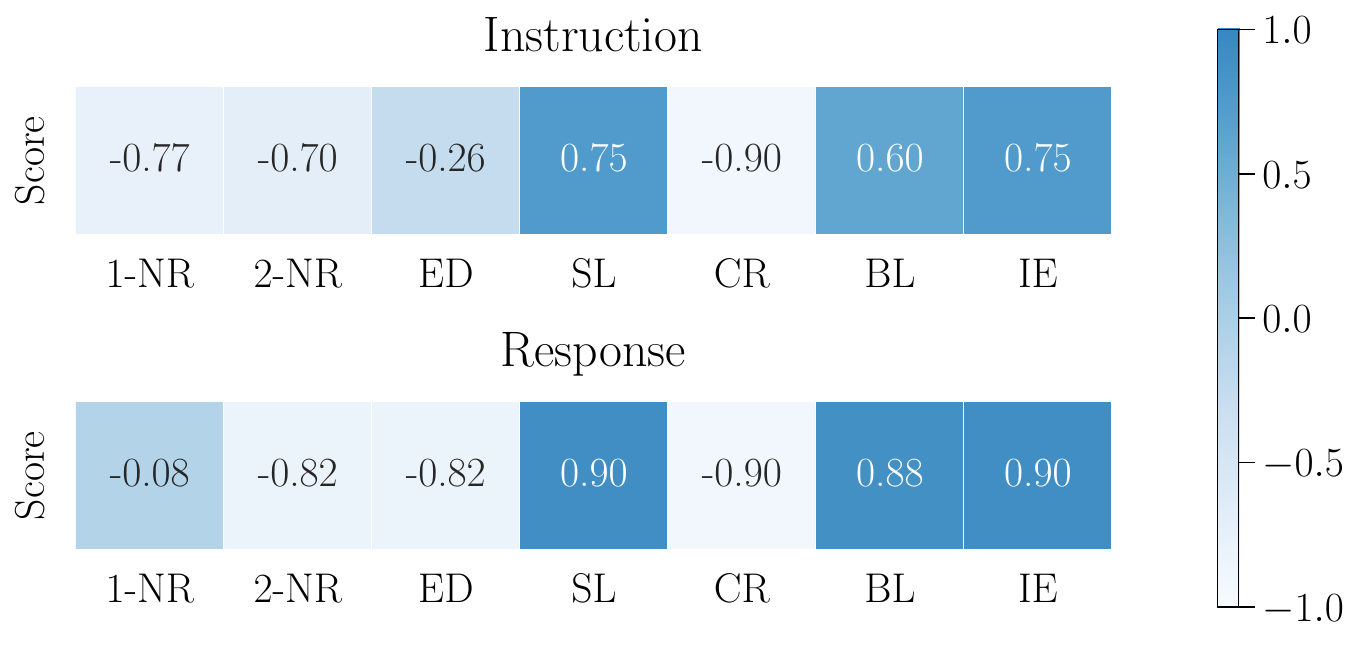}
\includegraphics[width=0.48\columnwidth]{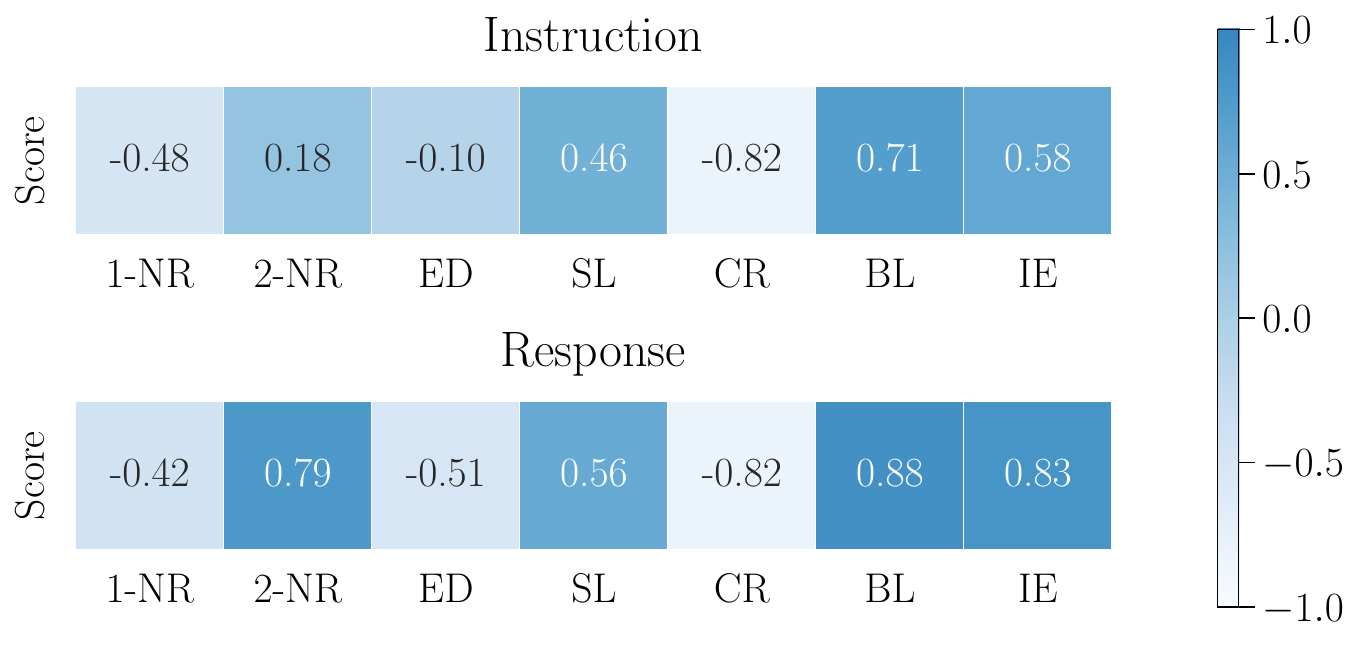}
\includegraphics[width=0.48\columnwidth]{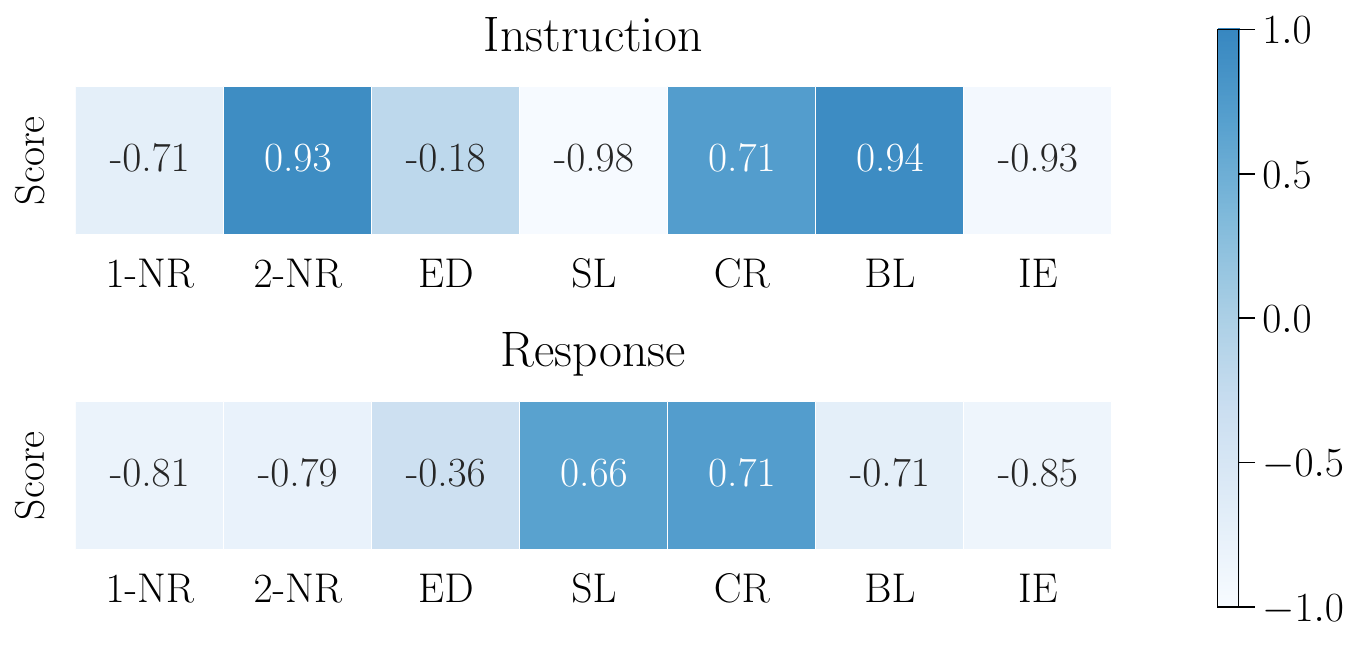}
\includegraphics[width=0.48\columnwidth]{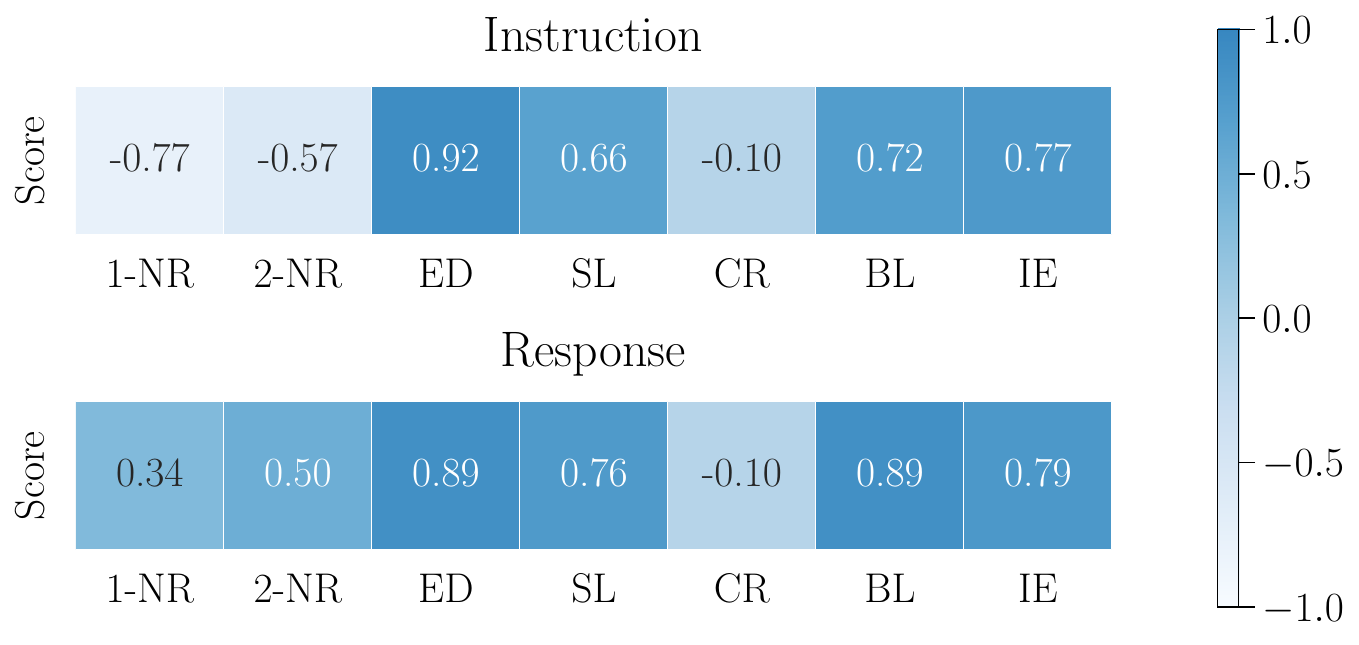}
\caption{Pearson correlation coefficients of multiple diversity parameters and model performance scores in the case of 10K dataset from different diversity-control strategies. The left panel shows the correlation analysis results based on the dataset constructed with instruction-based diversity-control strategies, while the right shows those from the response perspective. From top to bottom, the diversity-control strategies are at the macroscopic, mesoscopic, and microscopic levels, respectively. The settings and styles are the same as those in Figure~\ref{fig:relation_10K}. }
\label{fig:Instruction_response_relation}
\end{figure*}

\clearpage

\section{Prompt Template}
In this section, we present the prompt templates used for both the benchmark pairwise judgment task and the mesoscopic diversity strategy.

\subsection{Evaluate Prompt} 
In model evaluation, we adopt the pairwise methodology from Arena Hard~\cite{li2024crowdsourced} and evaluate on both the Arena Hard and AlpacaEval 2.0~\cite{alpaca_eval} benchmarks. Below is the pairwise judgment prompt template.

\label{app:judge_prompt}
\begin{tcolorbox}[
    enhanced,
    colframe=black,        
    colback=white,         
    coltitle=black,        
    colbacktitle=gray!30,  
    fonttitle=\bfseries,   
    title=Pairwise Judge Prompt Template, 
    rounded corners,       
    boxrule=0.5mm,         
    width=\textwidth,      
    left=2mm,              
    right=2mm,             
    top=2mm,               
    bottom=2mm,            
]

\texttt{system\_prompt: "Please act as an impartial judge and evaluate the quality of the responses provided by two AI assistants to the user prompt displayed below. You will be given assistant A's answer and assistant B's answer. Your job is to evaluate which assistant's answer is better.}

\texttt{Begin your evaluation by generating your own answer to the prompt. You must provide your answers before judging any answers.}

\texttt{When evaluating the assistants' answers, compare both assistants' answers with your answer. You must identify and correct any mistakes or inaccurate information.}

\texttt{Then consider if the assistant's answers are helpful, relevant, and concise. Helpful means the answer correctly responds to the prompt or follows the instructions. Note when user prompt has any ambiguity or more than one interpretation, it is more helpful and appropriate to ask for clarifications or more information from the user than providing an answer based on assumptions. Relevant means all parts of the response closely connect or are appropriate to what is being asked. Concise means the response is clear and not verbose or excessive.}

\texttt{Then consider the creativity and novelty of the assistant's answers when needed. Finally, identify any missing important information in the assistants' answers that would be beneficial to include when responding to the user prompt.}

\texttt{After providing your explanation, you must output only one of the following choices as your final verdict with a label:}

\texttt{1. Assistant A is significantly better: [[A>>B]]}

\texttt{2. Assistant A is slightly better: [[A>B]]}

\texttt{3. Tie, relatively the same: [[A=B]]}

\texttt{4. Assistant B is slightly better: [[B>A]]}

\texttt{5. Assistant B is significantly better: [[B>>A]]}

\texttt{Example output: \{My final verdict is tie: [[A=B]]\}}

\texttt{prompt\_template: ["<|User Prompt|> \{question\_1\}}

\texttt{<|The Start of Assistant A's Answer|>}

\texttt{\{answer\_1\}}

\texttt{<|The End of Assistant A's Answer|>}

\texttt{<|The Start of Assistant B's Answer|>}

\texttt{\{answer\_2\}}

\texttt{<|The End of Assistant B's Answer|>"]}

\end{tcolorbox}

\clearpage

\subsection{Tag Process Prompt} \label{app:tag_prompt}
In the mesoscopic diversity strategy, the prompt templates are adapted from Instag~\cite{lu2023instag}.
\subsubsection{Tagging Prompt} 

\begin{tcolorbox}[
    enhanced,
    colframe=black,        
    colback=white,         
    coltitle=black,        
    colbacktitle=gray!30,  
    fonttitle=\bfseries,   
    title=Tagging Prompt Template, 
    rounded corners,       
    boxrule=0.5mm,         
    width=\textwidth,      
    left=2mm,              
    right=2mm,             
    top=2mm,               
    bottom=2mm,            
]

\texttt{You are a tagging system that provides useful tags for instruction intentions to distinguish instructions for a helpful AI assistant. Below is an instruction:}

\texttt{[begin]}

\texttt{\{instruction\}}

\texttt{[end]}

\texttt{Please provide coarse-grained tags, such as "Spelling and Grammar Check" and "Cosplay", to identify main intentions of above instruction. Your answer should be a list including titles of tags and a brief explanation of each tag. Your response have to strictly follow this JSON format: [\{"tag":str, "explanation":str\}]. Please response in English.}

\end{tcolorbox}

\subsubsection{Tag Precision Evaluate Prompt}   
\label{app:filter_prompt}
\begin{tcolorbox}[
    enhanced,
    colframe=black,        
    colback=white,         
    coltitle=black,        
    colbacktitle=gray!30,  
    fonttitle=\bfseries,   
    title=Tag Precision Evaluate Template, 
    rounded corners,       
    boxrule=0.5mm,         
    width=\textwidth,      
    left=2mm,              
    right=2mm,             
    top=2mm,               
    bottom=2mm,            
]
\texttt{You are an experienced judge for intention tags of instructions. You will be provided a query and a list of tags describing intentions of the query as followed:}

\texttt{[query]:\{query\}}

\texttt{\{tags\}}

\texttt{Please provide feedback about whether all tags precisely describe an intention of the instruction. Please identify all incorrect tags and provide their indices in the JSON format as your response. The JSON format for your response is a list of JSON dictionary and the JSON dictionary has only one key to identify the index of each incorrect tag: [\{"idx":int\}]. For example, if [tag 0] and [tag 2] are incorrect, you should response as [\{"idx":0\}, \{"idx":2\}]. If all tags are correct, please response an empty list as [].}

\end{tcolorbox}

\subsubsection{Tag Consistency Evaluate Prompt}   
\label{app:consis_prompt}
\begin{tcolorbox}[
    enhanced,
    colframe=black,        
    colback=white,         
    coltitle=black,        
    colbacktitle=gray!30,  
    fonttitle=\bfseries,   
    title=Tag Consistency Evaluate Prompt Template, 
    rounded corners,       
    boxrule=0.5mm,         
    width=\textwidth,      
    left=2mm,              
    right=2mm,             
    top=2mm,               
    bottom=2mm,            
]
\texttt{You are an experienced judge for consistency of intention tags for instructions. You will be provided a tag and a list of instructions labeled with this tag as followed:}

\texttt{[tag]:\{tag\}}

\texttt{\{instructions\}}

\texttt{Please provide feedback about whether the meaning of this tag is consistent among all instructions. Please identify all inconsistent instructions and provide their indices in the JSON format as your response. The JSON format for your response is a list of JSON dictionary: [\{"idx":int\}]. For example, if the meaning of tags in [instruction 0] and [instruction 2] are inconsistent, you should response as [\{"idx":0\}, \{"idx":2\}]. If the meaning of tag is consistent in all instructions, please response an empty list as [].}

\end{tcolorbox}

\end{document}